\pdfoutput=1
\documentclass[table]{article}

\usepackage{colm2024_conference}
\usepackage{microtype}
\usepackage{hyperref}
\usepackage{url}
\usepackage{booktabs}
\usepackage{amsmath}
\usepackage{inconsolata}
\usepackage{graphicx, caption, subcaption}

\usepackage{amsmath,amsfonts,bm}

\def\eqref#1{equation~\ref{#1}}

\def\1{\bm{1}}

\DeclareMathAlphabet{\mathsfit}{\encodingdefault}{\sfdefault}{m}{sl}
\SetMathAlphabet{\mathsfit}{bold}{\encodingdefault}{\sfdefault}{bx}{n}

\usepackage{subcaption}
\definecolor{LightCyan}{rgb}{0.88,1,1}
\usepackage{tcolorbox} %
\tcbuselibrary{most}

\title{Rephrasing the Web:\\A Recipe for Compute and Data-Efficient Language Modeling}

\author{Pratyush Maini\thanks{Equal Contribution} \hspace{0.5mm}\thanks{Work done during internship at Apple} \\
Carnegie Mellon Univeristy\\
\texttt{pratyushmaini@cmu.edu}
\And Skyler Seto\footnotemark[1], He Bai, David Grangier, Yizhe Zhang, Navdeep Jaitly\\
Apple \\
\texttt{\{sseto,hbai22,grangier,yizhe\_zhang,njaitly\}@apple.com}\\
}

\newcommand{\ours}{\textbf{WRAP}}

\colmfinalcopy %
\begin{document}

\maketitle

\begin{abstract}
    Large language models are trained on massive scrapes of the web, which are often unstructured, noisy, and poorly phrased. 
    Current scaling laws show that learning from such data requires an abundance of both compute and data, which grows with the size of the model being trained. 
     This is infeasible both because of the large compute costs and duration associated with pre-training, and the impending scarcity of high-quality data on the web.  
    In this work, we propose \textbf{W}eb \textbf{R}ephrase \textbf{A}ugmented \textbf{P}re-training~(\ours) that uses an off-the-shelf instruction-tuned model prompted to paraphrase documents on the web in specific styles such as ``like Wikipedia'' or in ``question-answer format'' to jointly pre-train LLMs on real and synthetic rephrases. First,  we show that using \ours~on the C4 dataset, which is naturally noisy, 
    speeds up pre-training by $\sim3\times$. At the same pre-training compute budget, it
    improves perplexity by more than 10\% on average across different subsets of the Pile, and improves zero-shot question answer accuracy across 13 tasks by more than 2\%. Second, we investigate the impact of the re-phrasing style on the performance of the model, offering insights into how the composition of the training data can impact the performance of LLMs in OOD settings. Our gains are attributed to the fact that re-phrased synthetic data has higher utility than just real data because it (i) incorporates style diversity that closely reflects downstream evaluation style, and (ii) has higher `quality' than web-scraped data.

\end{abstract}

\section{Introduction}
\label{sec:intoduction}
Large language model (LLM) pre-training has been largely democratized and open-sourced, allowing various academic labs, and industries to pre-train custom LLMs. 
Yet, a key differentiator between these models is the composition and size of the data used to train them. 
Data curation strategies are required to filter out scrapes of the web that are unstructured and/or poorly phrased~\citep{eisenstein-2013-bad}.
While some of these strategies have been made public~\citep{brown2020language,wenzek-etal-2020-ccnet,penedo2023refinedweb}, most state-of-the-art data curation techniques are unknown to the research community, and only anecdotal evidence remains. Research on data curation requires multiple rounds of re-training, making it an expensive endeavour to document techniques that lead to practical improvements.
On the other hand, scaling laws for language models (such as Chinchilla scaling laws~\citep{hoffmann2022training}) show that with increasing model sizes, we should also increase both the training compute and data size linearly.  This is infeasible because (a) high-quality data is limited \citep{villalobos2022will}, and repeating for even a small number of epochs (4 or more) results in diminishing returns or overfitting \citep{muennighoff2023scaling,touvron2023llama,xue2023repeat}; and (b) pre-training for such long durations is prohibitively expensive. 

Meanwhile, the use of synthetic data has gained prominence in the paradigm of aligning pre-trained LLMs via instruction fine-tuning, RLHF~\citep{ouyang2022training}, and instruction backtranslation \citep{li2023self}. 
Recently, in the context of pre-training, synthetic data was used to generate datasets such as Tiny Stories~\citep{eldan2023tinystories}
and Textbook quality synthetic data~\citep{gunasekar2023textbooks,li2023textbooks}.  These were used to train smaller language models (like the Phi model family) that were as performant as larger language models on certain tasks.  
However, their data generation process stays largely opaque, and prohibitively expensive, requiring prompting a GPT-3.5 model for generating billions of tokens. Additionally, such data generation can create a large ``knowledge bias'' by specifically generating data pertaining to tasks that we want to perform well on. While synthetic data has shown promise, it is unclear if this is because of the higher quality nature of synthetic data, or because of strategic topic selection~\citep{maini_phi_1_5}.
\begin{figure}[t]
    \centering
       \begin{subfigure}{0.32\textwidth}
        \centering
        \includegraphics[width=\textwidth]{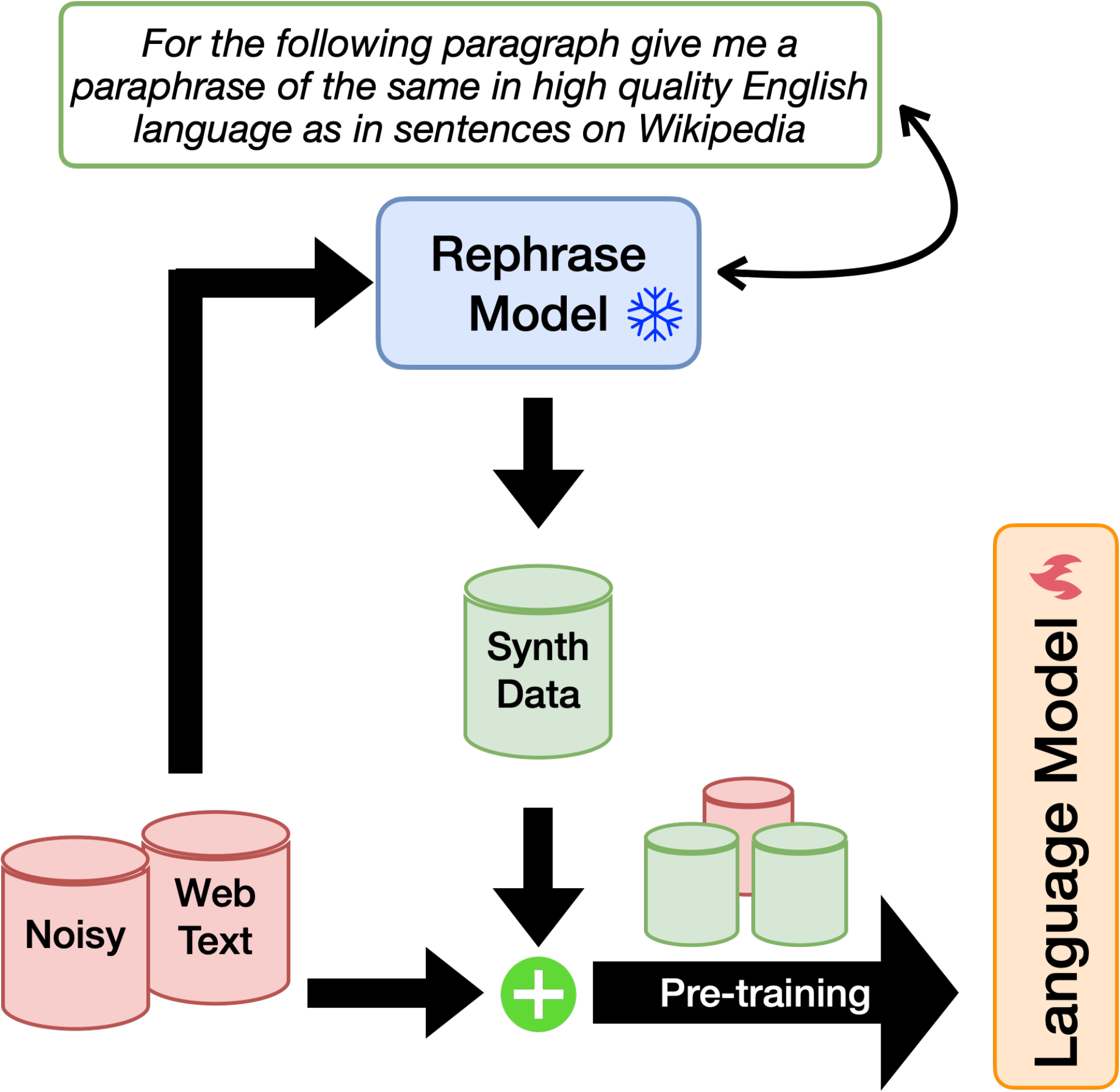}
        \caption{}
        \label{fig:method}
        \vspace{2mm}
    \end{subfigure}
    \hfill
    \begin{subfigure}{0.32\textwidth}
        \centering
        \includegraphics[width=\textwidth]{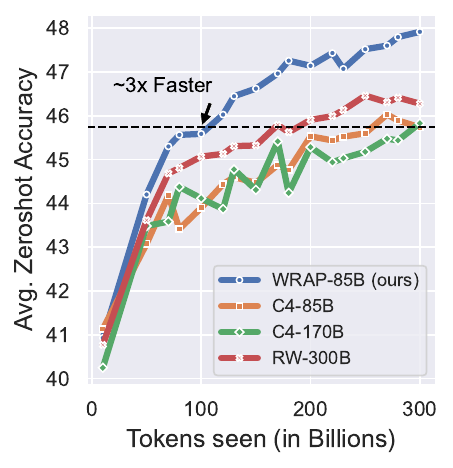}
        \caption{}
        \label{fig:upper_right_image}
    \end{subfigure}
    \hfill
    \begin{subfigure}{0.32\textwidth}
        \centering
        \includegraphics[width=\textwidth]{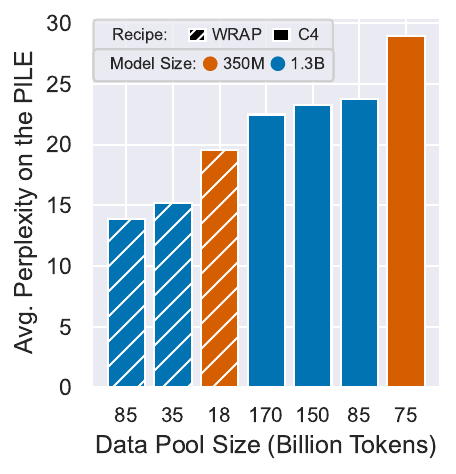}
        \caption{}
        \label{fig:lower_right_image}
    \end{subfigure}

    \caption{(a) \ours~Recipe: We prompt an off-the-shelf instruction-tuned model to rephrase articles on the web, and pre-train an LLM on a mixture of real and synthetic data. ~(b) Zero-shot performance of GPT 1.3B models trained on combinations of C4 and synthetic variations. Each step corresponds to a batch of 1M samples. (c) Weighted average perplexity over 21 sub-domains of the Pile for varying model sizes and amount of pre-training data.}
    \label{fig:common_caption}
\end{figure}

In this work, we propose \textbf{W}eb \textbf{R}ephrase \textbf{A}ugmented \text{P}re-training~(\ours)---that attempts to bridge
three important challenges stemming from the ambiguity around data curation---
(i) what data should you pre-train on?
(ii) how can you pre-train with limited data?
(iii) how can you pre-train computationally efficiently?
In particular, we show that re-phrasing documents on the web using an off-the-shelf medium size LLM allows models to learn much more efficiently than learning from raw text on the web, and 
accounts for 
performance gains on out of distribution datasets that \emph{can not} be offset with additional web data. Our proposed method involves using a pre-trained off-the-shelf LLM to re-phrase documents from a web corpus into different styles.  An overview of our approach is shown in Figure~\ref{fig:method}.

In our work, we tackle two important challenges faced during synthetic data curation in the works of~\citet{gunasekar2023textbooks}---generation cost, and data bias---by rephrasing articles on the web. (i) \ours~allows for using an open source, and much smaller LLM (1.8B/7B v/s GPT3.5) to rephrase unstructured and poorly phrased documents in different styles, since it does not rely on the LLM as a knowledge bank. (ii) Thanks to the information maintaining nature of rephrasing, we are able to leverage the natural diversity of the web, rather than relying on an LLM for information which may be prone to factual errors, and/or data biases. Our work shows that the ``style'' alone can result in large gains in downstream performance.

Using \ours~on the C4, we evaluate model performance on 13 different zero-shot tasks, and 21 different language modeling domains of the Pile,  and find that pre-training LLMs with synthetic data allows us to train equivalent models with 5x lesser data, or 3x lesser compute. In fact, our synthetic data trained models, also outperform the recent TinyLLama models that were trained for 3 trillion tokens (10x data and compute) across several zero-shot Q/A tasks. 
We further observe 
a reduction in perplexity by $\sim50\%$ on the Pile,
and note that our 350M parameter model trained on combinations of real and synthetic rephrases on just $15\%$ of the entire C4 corpus,  outperforms pre-training a 1.3B parameter on the entire C4.
Finally, we conduct an analysis on the 
potential of data leakage, 
properties of synthetic data styles, and how to combine synthetic data for improving  \ours~based LLM pre-training.

\section{Related Work}
\paragraph{Neural Scaling Laws for Language Models}
Neural scaling laws relate the optimal number of model parameters and amount of training data for a fixed amount of compute.  \citet{hoffmann2022training} presented the Chinchilla scaling laws for language models demonstrating that there was a linear relationship between the size of the model and the amount of training data needed. Their findings indicated that prior models such as Gopher~\citep{rae2021scaling} are severely undertrained. Recently, models such as Llama~\citep{touvron2023llama} are trained with much more data.  These scaling laws were drawn for the paradigm of single-epoch training.  Recently, \citet{muennighoff2023scaling} showed that
the marginal utility of repeated data rapidly diminishes when training for more than 4 epochs, and formulated scaling laws under repeated data.
Concurrently, \citet{xue2023repeat} showed that repeating even small fractions of the pre-training data can lead to overfitting and reduce model performance.

\paragraph{Dataset Selection}
Selecting high quality data for pre-training LLMs remains an active, high-impact, yet understudied area of research. For instance,
 GPT-2 model was pre-trained on all outbound links from Reddit, a social media platform, which received at least 3 karma~\citep{brown2020language}. This was used as a heuristic indicator that the document may be \emph{interesting, educational, or just funny.}
 Follow-up works have used other heuristics such as prioritizing documents that resemble wikipedia~\citep{gururangan-etal-2022-whose}.~\citet{rae2021scaling}
used multiple heuristic filters to remove documents, such as the absence of certain stopwords, length of the document, percentage of alphabetic characters, 
 mean word length, 
  symbol-to-word ratio, percentage of lines starting with a bullet point, or ending with an ellipsis etc. Their work highlights the intricacies of filtering out text data.
An alternative paradigm for building better datasets for training is to distill high-quality datasets.  \citet{xie2023doremi} proposed a method, DoReMi, to select the best data mixture for pre-training language models by reweighting data from various domains.
Concurrently, \citet{Abbas2023SemDeDupDL} showed that de-duplicating pre-training data can improve pre-training efficiency.
Recently several methods were proposed for automatic filtering of low-quality data for faster fine-tuning of LLMs~\citep{chen2023alpagasus,solaiman2021process,zhou2023lima}. Simultaneously, in the realm of image-language models such as CLIP~\citep{radford2021learning}, the Datacomp benchmark~\citep{gadre2023datacomp} and recent entries~\citep{maini2023t,yu2023devil} have developed approaches at filtering out low-quality subsets from pre-training datasets like LAION~\citep{schuhmann2022laion}, or from scrapes of the common crawl.

\paragraph{Data Augmentation and synthetic data}
\citet{eldan2023tinystories} showed that a synthetically generated dataset in the form of stories that toddlers can understand
allows training a small language model that can generate coherent sentences.
\citet{gunasekar2023textbooks} showed that textbook quality (synthetic) data alone helps models achieve state-of-the-art performance on reasoning and coding tasks. Similar approaches are used in concurrent work for enhancing coding and mathematical reasoning abilities while finetuning \cite{liu2023tinygsm,wei2023magicoder}.
\citet{shumailov2023model} show that training on synthetic data can actually be harmful for model performance, especially when
we do multiple rounds of pre-training an LLM and then training the next LLM on data generated by the previous one.
On the other hand, some other works have shown that such a strategy can actually be useful.
\citet{li2023self} and \citet{koksal2023longform} discuss how a model can generate instruction data and then fine-tune on its own generated data
to improve performance.
\citet{jung2023impossible} discuss how such repeated cycles of synthetic data can help train a very small paraphrase and summarization model that even outperforms GPT-3.

The vision and multimodal literatures have also seen a surge of works examining the use of synthetic data for training.
The works of \citet{bansal2023leaving,trabucco2023effective,azizi2023synthetic} have shown that using synthetic data in 
combination with real data achieves state of art model performance both in-distribution and out-of-distribution. \citet{cubuk2020randaugment} used generative models to generate image augmentations for better domain generalization. There are also multiple studies on increasing multiplicity of augmentations and their value for improving generalization \citep{choi2019faster,fort2021drawing,hoffer2020augment}.  
However, \citet{alemohammad2023self} showed that generated models trained for more than five cycles of their own generated data can undergo severe mode collapse.

\section{\ours: Web Rephrase Augmented Pretraining}
Generating synthetic data using an off-the-shelf language model can be both computationally expensive and operationally challenging. Prior approaches to generating synthetic textbook quality data using LLMs~\citep{gunasekar2023textbooks} required (1) a language model that contains sufficient world knowledge to generate articles worth training on, thereby increasing generation cost; (2) a careful selection of prompts that enable generating high quality, and diverse articles that fill any knowledge gaps in the synthetic corpus. This challenge was highlighted in follow-up work of \citet{li2023textbooks}, and has the potential of inadvertently creeping in biases in the language models~\citep{maini_phi_1_5}, as opposed to those trained on the natural diversity of the web.
As a remedy to the challenge of (i) generation cost, and (ii) data diversity, we propose \ours~ that leverages the natural diversity of articles on the web, allowing us to utilize significantly smaller LLMs (than GPT-3.5) to generate high-quality paraphrases of noisy and unstructured articles on the web.

\subsection{Rephrasing the Web}
\label{sec:synthetic_data_generation}
It has been observed in past work that up-weighting high-quality data, such as texts from Wikipedia, can be useful to improve language modeling. These terms have generally been very loosely defined and there is only anecdotal evidence of the same~\citep{brown2020language,wenzek-etal-2020-ccnet}. At the same time, web data is deficient of text in question-answering or conversational format, which is a prominent use case of language models. Based on these two insights, we design the rephrasing styles for our work.

\paragraph{Rephrasing Styles}
In lieu of the anecdotal evidence above, we attempt rephrasing documents on the web in four different styles---(i) Easy (text that even a toddler will understand); (ii) Medium (in high quality English such as that found on Wikipedia); (iii) Hard (in terse and abstruse language); (iv) Q/A (in conversation question-answering format). In order to operationalize rephrasing in these stylistic variations, we appropriately prompt an instruction-tuned model. The rephrased examples of these four styles and the prompts templates used in our work are provided in Appendix~\ref{app:prompt_styles}.

\paragraph{Generating Synthetic Data}
Now, we detail how we utilize an instruction-tuned language model to rephrase texts from web-crawled datasets such as C4~\citep{raffel2020exploring} (which we use for all our experiments). In particular, we use a frozen Mistral-7B instruction-tuned model~\citep{jiang2023mistral} (see Ablations in Section~\ref{sec:ablations} for other models).
To generate synthetic data in ``medium'' style, the Mistral model is prompted using the following instruction: \emph{``For the following paragraph give me a paraphrase of the same in high-quality English language as in sentences on Wikipedia''}. The prompt was created using iterative human feedback by comparing outputs of `medium' sized LLMs with those of GPT-4.
We use the model output to create a parallel corpus of ``high-quality'' synthetic data corresponding to the original noisy web data. Each example has a maximum of 300 tokens, which was decided based on our empirical observation that asking an LLM to rephrase more than 300 tokens, often led to loss of information. Discussions on data quality can be found in Section~\ref{sec:synth_properties}.

\paragraph{Combining Real and Synthetic Data}
Our method of re-phrasing web data naturally incorporates the information diversity found on the internet. However, it does not incorporate the noise 
in real data. While synthetic data may help LLMs pre-train faster, we also want them to be able to understand noisy web text that may be filled with typos and linguistic errors so that the LLMs do not fail in user facing situations. In order to incorporate this style diversity in language modeling, we sample real and synthetic data in a 1:1 ratio.

\subsection{Implementation Details}
\label{sec:implementation_details}
\paragraph{Architecture}
We train decoder-only transformer models ~\citep{vaswani2017attention} at three different scales, small, medium and XL. 
The small-scale (128M parameter) model consists of 12 layers, 12 attention heads, and a hidden dimension size of 768.
The medium-scale (350M parameter) model consists of 24 layers, 16 attention heads, and a hidden dimension size of 1024.
The XL-scale (1.3B parameter) model consists of 24 layers, 16 attention heads, and a hidden dimension size of 2048.
We do not use dropout in either model and a maximum sequence length of 1024. The models are trained using NVIDIA's \href{https://github.com/NVIDIA/Megatron-LM}{Megatron-LM} repository.

\paragraph{Pre-training}
We train all our XL models for a total of 300k steps with a batch size of one million tokens, unless otherwise specified.
We use a maximum learning rate of $3e^{-4}$ for the 128M, and 350M parameter models, and $2e^{-4}$ for the 1.3B parameter model. The minimum learning rate is $1e^{-5}$.  We use a weight decay of $0.01$, along with a gradient clipping norm of $1.0$.
We use cosine learning rate scheduler with a warmup for 1\% of the total steps;
 and the Adam optimizer with $\beta_1=0.9$ and $\beta_2=0.999$.

\begin{figure}[t]
\centering
\includegraphics[width=\linewidth]{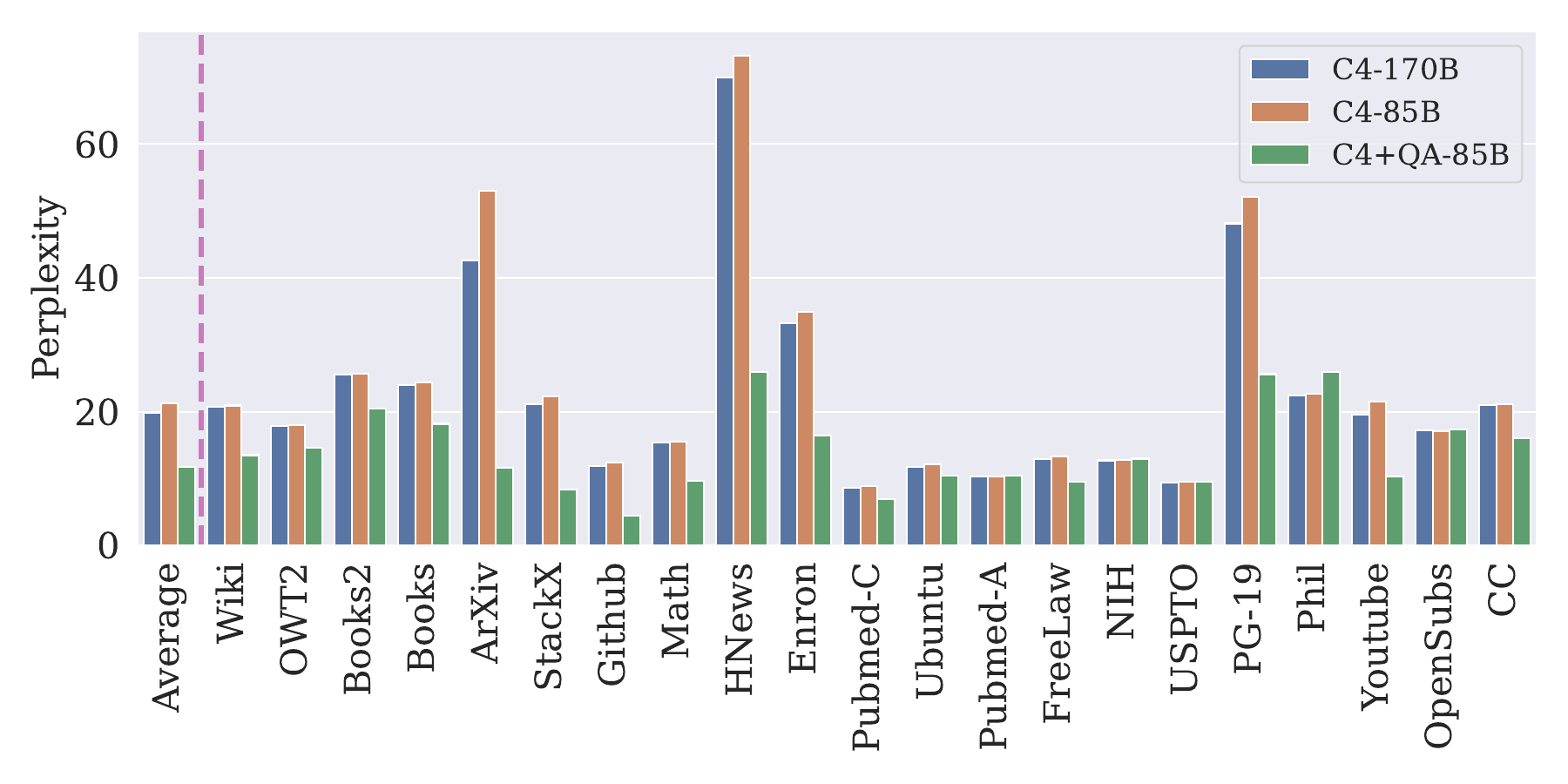}
\caption{\textbf{\ours~(C4 + QA-85B) v/s C4}: Comparison of perplexity on the Pile  for a 1.3B LLM trained for 300B tokens shows that WRAP outperforms models trained on 2x real data.}
\label{fig:pile_wavg}
\end{figure}

\section{Perplexity Evaluation}
\label{sec:results}

We evaluate the perplexity of the pre-trained model on the validation set of multiple out-of-distribution datasets.  
All models are either trained on the C4 dataset~\citep{raffel2020exploring}, or a particular stylistic rephrase of the same. All the evaluations are done on 21 sub-domains of the Pile~\citep{gao2020pile}.
These subsets
are created from the first 10,000 documents from each domain of the Pile dataset. We then evaluate the perplexity of the model on these subsets.  Additional evaluation details are provided in Appendix~\ref{app_eval}.
It is important to note that we evaluate perplexities on the Pile instead of C4.
Training on multiple distributions of text (synthetic and real web) does come at a small cost of less than 1 perplexity on the C4 validation set.  To understand our choice of evaluation, and why we observe this perplexity increase, we note that training over the C4 corpus corresponds to minimizing the objective 
\begin{equation}
    \theta_{c4}   = \min_{\theta} \mathbb{E}_{x \sim D_{c4}}\left[ \mathcal{L}(\theta; x) \right], 
    \label{eq:theta}
\end{equation}
that attempts to exactly model the C4 web text.  In contrast, training over multiple styles corresponds to minimizing the risk over a different distribution,

\begin{equation}
    \theta_{\ours}   = \min_{\theta} \mathbb{E}_{x \sim D_{c4}\cup D_{syn}}\left[ \mathcal{L}(\theta; x) \right] .
    \label{eq:theta_syn}
\end{equation}

Solving for \eqref{eq:theta_syn} does not minimize the risk over C4-only, and hence it is unfair to compare
$\theta_{c4}$ and
$\theta_{\ours}$ on the C4.
For meaningfully comparing models trained on the C4 and on its synthetic rephrases, we evaluate their generalization capability on 21 different domains of the Pile~\citep{gao2020pile}. 
Results for each domain are presented in Figure~\ref{fig:pile_wavg}.

\paragraph{Data Complexity}
In Figure~\ref{fig:lower_right_image}, we show that models trained for fewer tokens (150B) and even smaller 350M models outperform training on the full C4 for 300B tokens indicating faster learning with synthetic rephrases.  
On some domains such as ArXiv and HackerNews, we observe that training with synthetic data allows reducing the perplexity by nearly 3x of the perplexity of models trained on real data alone. This suggests that in many cases 
it may not be possible to offset 
the performance advantage of pre-training on synthetic data
by merely training on more real data.
Overall, on an average of multiple subsets of the Pile,
our models improve perplexity by 50\% over models trained on real data alone. 

\paragraph{Learning Speed}
We observe that even at the first checkpoint (10B tokens) of \ours~training, the average perplexity of the LLM on the Pile is lower than that achieved by pre-training on C4 for 15 checkpoints. This suggests a 15x pre-training speed-up. We defer the discussion on learning speed to  `zero-shot' tasks in order to make more meaningful comparisons.

\section{Zero-shot Tasks}
We now evaluate our pre-trained language models on various zero-shot question answering (QA) benchmarks
using the LLM Evaluation Harness\footnote{We use git commit - \texttt{89618bf8} for consistency across all experiments with a batch size of 32.}~\citep{eval-harness}.

\subsection{Datasets}
We evaluate our models on a total of 13 different zero-shot benchmarks to assess their abilities across various natural language tasks like common sense reasoning, language and knowledge understanding and mathematical reasoning.

\paragraph{General Understanding}
The General Understanding category comprises datasets testing broader cognitive skills and language comprehension. \textbf{ARC Easy (ARC-E)}~\citep{allenai:arc} is the less challenging counterpart of ARC-C, featuring questions that require basic reasoning skills. \textbf{BoolQ}~\citep{clark-etal-2019-boolq} includes boolean questions that focus on reading comprehension and general language understanding. \textbf{Winogrande (Wino.)}~\citep{ai2:winogrande} challenges models with common sense reasoning in language, particularly in pronoun disambiguation. \textbf{PIQA}~\citep{Bisk2020} assesses understanding of physical processes, an essential part of practical common sense. \textbf{HellaSwag}~\citep{zellers2019hellaswag} tests the ability to complete scenarios coherently, demanding both language understanding and common sense. \textbf{TruthfulQA}~\citep{lin2021truthfulqa} is centered on generating truthful, accurate answers, thus testing the model's factual correctness. \textbf{OpenBookQA (OBQA)}~\citep{OpenBookQA2018} evaluates the understanding of a broad range of facts and concepts. Finally, \textbf{LogiQA-2}~\citep{logiqa2} assesses the model's capacity to comprehend and apply logical principles.

\paragraph{Specialized Knowledge}
In the Specialized Knowledge category, we include datasets that demand expertise in specific domains. The \textbf{ARC Challenge (ARC-C)}~\citep{allenai:arc} contains challenging science exam questions from grades 3 to 9, demanding advanced knowledge. \textbf{SciQ}~\citep{SciQ} provides science exam questions to test models' understanding and reasoning in the scientific domain. \textbf{PubMedQA}~\citep{jin2019pubmedqa} focuses on biomedical literature, assessing comprehension in medical and health-related information. 
\textbf{MathQA}~\citep{amini-etal-2019-mathqa} tests mathematical problem-solving, requiring both numerical comprehension and reasoning. Lastly, \textbf{MMLU}~\citep{hendryckstest2021} spans multiple domains, from professional to academic, testing the model on specialized subjects.

\begin{table*}[t]
    \centering
    \scalebox{0.8}{
    \begin{tabular}{lcccccccc>{\columncolor{LightCyan}}c}
        \toprule
        Dataset (Real Tok.) & ARC-E & BoolQ & Wino. & PIQA & HellaSwag & TruthfulQA & OBQA & LogiQA & Avg \\
        \midrule
        Half C4 (85B) & 61.2  &  59.1  &  57.3  &  74.9  &  46.5  &  34.1  &  22.4  &  23.5  &  47.4 \\
        Full C4 (170B) & 61.6  &  54.2  &  59.0  &  74.9  &  46.8  &  33.5  &  25.0  &  23.4  &  47.3 \\
        RW (160B) &  61.6  &  60.7  &  57.5  &  74.3  &  45.2  &  36.8  &  21.8  &  23.2  &  47.6 \\
        RW (320B) & 60.7  &  61.1  &  57.1  &  74.4  &  45.6  &  36.0  &  22.6  &  22.5  &  47.5 \\
        Pythia-Pile (300B) &   60.5  &  63.3  &  57.5  &  70.8  &  40.4  &  38.9  &  22.2  &  22.2  &  47.0 \\
        TinyLlama (1T) & 60.3  &  57.8  &  59.1  &  73.3  &  45.0  &  37.6  &  21.8  &  24.5  &  47.4 \\
        \midrule
        Synthetic (85B) &63.9  &  60.0  &  58.8  &  76.1  &  45.2  &  44.0  &  23.0  &  24.1  &  49.4\\
        Synthetic+C4 (85B) & 64.1  &  62.2  &  58.9  &  75.4  &  46.2 &  40.6  &  24.1  &  23.9  &  49.4 \\
        \bottomrule
    \end{tabular}}
    \caption{Evaluation of $\sim1.3$B parameter LLMs on `General Understanding Tasks'  on datasets focusing on general reasoning, language understanding, and common sense. Results for \ours are averaged over 3 runs}
    \label{tab:general_understanding}
\end{table*}

\begin{table*}[t]
    \centering
    \scalebox{0.8}{
    \begin{tabular}{lccccc>{\columncolor{LightCyan}}c}
        \toprule
        Dataset (Real Tok.) & ARC-C & SciQ & PubMedQA & MathQA & MMLU & Avg \\
        \midrule
        Half C4 (85B) & 26.3  &  84.5  &  57.2  &  23.4  &  24.2  &  43.1 \\
        Full C4 (170B) & 26.8  &  85.0  &  57.4  &  24.3  &  23.9  &  43.5 \\
        RW (160B) & 27.2  &  87.2  &  56.2  &  24.1  &  25.9  &  44.1 \\
        RW (320B) & 27.8  &  88.0  &  57.4  &  23.0  &  25.4  &  44.3 \\
        Pythia-Pile (300B) &  26.1  &  86.6  &  60.6  &  25.2  &  24.3  &  44.6 \\
        TinyLlama (1T) & 27.8  &  88.9  &  61.4  &  24.1  &  25.8  &  45.6 \\
        \midrule
        Synthetic (85B) & 29.7  &  87.0  &  60.2  &  23.4  &  24.6  &  45.0 \\
        Synthetic+C4 (85B) & 29.9  &  87.6  &  61.5  &  23.9  &  24.8  &  45.5 \\
        \bottomrule
    \end{tabular}}
    \caption{Evaluation of $\sim1.3$B parameter LLMs on `Specialized Knowledge Tasks'  that require specific domain knowledge such as science, medicine, mathematics, and logic. Results for \ours are averaged over 3 runs.}
    \label{tab:specialized_knowledge}
\end{table*}

\subsection{Results}
We compare the performance of a model trained on a mixture of real and synthetic data with models trained on various splits of real data. In all our experiments, we use the C4~\citep{raffel2020exploring} dataset for rephrasing and producing splits of synthetic data. 
We use the abbreviation `Real Tok.' to denote the number of tokens of web data available for pre-training. In the `Synthetic + Real' experiments, we augment the same number of synthetic rephrases. We choose `Real Tokens' as the metric of comparison because we can potentially rephrase the same document multiple times, implying that the total corpus size is not meaningful, and corpus `knowledge' is the actual currency of interest.

\paragraph{Baselines Methods}
We pre-train LLMs of 
(i) Half of C4, and the (ii) Full C4 corresponding to approximately
85 Billion and 170 Billion real tokens respectively~\citep{raffel2020exploring}. We also pre-train our own models on (iii) 160 Billion and (iv) 320 Billion tokens of the RefinedWeb Dataset~\citep{penedo2023refinedweb}. Additionally, we also compare with the (iv) Pythia-1.4B model that was trained on the Pile~\citep{gao2020pile}. This dataset is no longer publicly available, hence we utilize a pre-trained model. Finally, we also compare with the recent (v) TinyLlama model~\citep{zhang2024tinyllama} that was trained for 3 epochs  on 1 Trillion tokens of data from 
SlimPajama \citep{shen2023slimpajama} and StarCoder \citep{li2023starcoder}.

\paragraph{General Improvements}
Across all tasks in Table~\ref{tab:general_understanding}, we observe that models trained on synthetic data combined with the C4 dataset (Synthetic+C4) exhibit an overall higher average performance of 49.4\% as compared to those trained solely on the real C4 dataset with a 85B token split, which scored an average of 47.4\%. This shows that the inclusion of synthetic data can enhance the general understanding capabilities of NLP models. Moreover, even the TinyLLama model trained for 10x compute and data, performs comparably to the other models trained on real data. This suggests that the gains from filtering out, or adding more real data are very low. As opposed to this, \ours~shows that pre-training on even small amounts of synthetic data can contribute to large performance gains.

\paragraph{Specialized Knowledge Tasks} 
The key message from the results
in Table~\ref{tab:specialized_knowledge} is that synthetic data can not impart `new knowledge'. It can only help pre-train faster, which was also the premise of our work. In particular, we note several key findings:
\begin{enumerate}
    \item Pre-training on larger datasets helps improve performance, by presumably exposing the LLM to more ``knowledge''. For instance, the Pythia (300B) model achieves an average score of 44.6\%, outperforming the smaller C4 (85B) dataset's score of 43.5\%.
    \item Despite the advantages of a larger dataset, the improvements saturate. For example, RefinedWeb (320B) model outperforms the RefinedWeb (160B) model by only 0.2\%. Similarly, the TinyLlama model (1T tokens) performs comparably to the \ours~ model, which only had 85B tokens of raw web data.
\end{enumerate}

\paragraph{Specific Improvements} 
We see maximum improvement in the TruthfulQA dataset, with the Synthetic (85B) model scoring 44.0\%, which is significantly higher than any other model's performance on this dataset. This is potentially because instruction-tuned LLMs already correct potential misconceptions while rephrasing the text. Interestingly, we notice how adding real data to the synthetic model (Synthetic+C4) reduces the performance on TruthfulQA by 4\%, down to 40.5\%, indicating a potential dilution of the benefits gained from synthetic data when combined with real data. Other datasets such as HellaSwag, and BoolQ, for which C4 trained models do well, continue to show the benefits of incorporating combinations of C4 and synthetic rephrases.

\section{Analysis and Ablations}
\label{sec:ablations}

We further ask the following Research Questions (RQs) to investigate in a finer granularity how to enhance performance optimally.

\subsection{Data Combination Analysis}

\paragraph{\textbf{RQ1: How important is it to have real C4 data?}}

\begin{figure}[t]
\centering
\includegraphics[width=\linewidth]{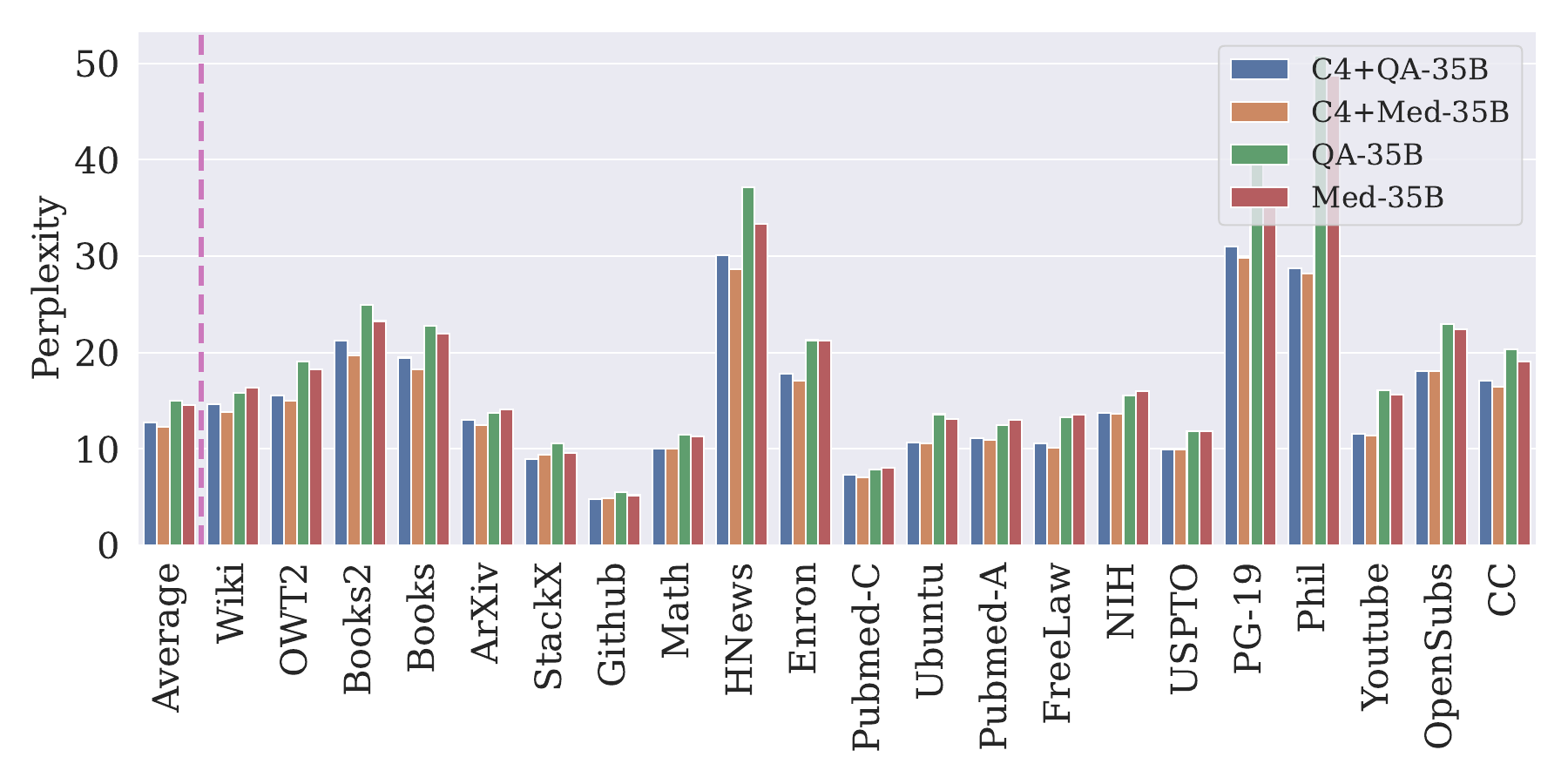}

\caption{\textbf{Importance of Real Data:} Comparing perplexity on the Pile when pre-training on C4 with synthetic data vs. synthetic data only.  Models are 1.3B parameters trained for a total of 150B tokens on a real data subset containing 35 Billion tokens of the C4.}
\label{fig:syn_comp}
\end{figure}

\begin{table*}[t]
    \centering
    \scalebox{0.8}{
    \begin{tabular}{lcccccccc>{\columncolor{LightCyan}}c}
        \toprule
        Dataset (Real Tok.) & ARC-E & BoolQ & Wino. & PIQA & HellaSwag & TruthfulQA & OBQA & LogiQA & Avg \\
        \midrule
        Med+C4-35B  & 59.8  &  57.0  &  55.7  &  74.6  &  44.5  &  36.5  &  23.8  &  21.5  &  46.7\\
        QA+C4-35B &  62.2  &  63.3  &  55.7  &  74.8  &  44.6  &  41.4  &  22.4  &  23.2  &  48.4\\
        Med-35B &  56.6  &  59.5  &  53.4  &  74.0  &  41.9  &  36.3  &  22.2  &  22.7  &  45.8\\
        QA-35B &  61.7  &  62.0  &  53.9  &  75.2  &  43.4  &  43.0  &  22.8  &  23.4  &  48.2\\
        \bottomrule
    \end{tabular}}
    \caption{\textbf{Importance of Real Data:} Evaluation of $\sim1.3$B parameter LLMs trained for 150B tokens on General Understanding Tasks. Results show that adding real data helps improve model performance when pre-training on `Medium' or `Wikipedia-style' paraphrases. 
    }
    \label{tab:general_understanding_150B_real_data}
\end{table*}

\begin{table*}[t]
    \centering
    \scalebox{0.8}{
    \begin{tabular}{lccccc>{\columncolor{LightCyan}}c}
        \toprule
        Dataset (Real Tok.) & ARC-C & SciQ & PubMedQA & MathQA & MMLU & Avg \\
        \midrule
       Med+C4-35B & 27.2  &  82.2  &  46.2  &  23.1  &  25.2  &  40.8\\
        QA+C4-35B & 29.0  &  85.1  &  62.2  &  22.5  &  26.1  &  45.0\\
        Med-35B & 27.0  &  80.0  &  59.4  &  22.5  &  24.7  &  42.7\\
        QA-35B &  27.1  &  85.5  &  59.2  &  22.2  &  25.0  &  43.8\\
        \bottomrule
    \end{tabular}}
    \caption{\textbf{Importance of Real Data:} Evaluation of $\sim1.3$B parameter LLMs on Specialized Knowledge Tasks.
    Results show that adding real data helps improve model performance when pre-training on `Q/A-style' paraphrases. }
    \label{tab:specialized_knowledge_real_data}
\end{table*}

\begin{table*}[t]
    \centering
    \scalebox{0.8}{
    \begin{tabular}{lccccc>{\columncolor{LightCyan}}c}
        \toprule
        Dataset (Real Tok.) & ARC-C & SciQ & PubMedQA & MathQA & MMLU & Avg \\
        \midrule
       Med+C4-35B & 27.2  &  82.2  &  46.2  &  23.1  &  25.2  &  40.8\\
        QA+C4-35B & 29.0  &  85.1  &  62.2  &  22.5  &  26.1  &  45.0\\
        Combined-1:1-35B &  28.2  &  85.9  &  61.2  &  23.2  &  23.9  &  44.5 \\
        Combined-1:2-35B &  29.0  &  85.7  &  57.4  &  23.5  &  23.1  &  43.7\\
        \bottomrule
    \end{tabular}}
    \caption{\textbf{Combining multiple styles:} Evaluation of $\sim1.3$B parameter LLMs trained for 150B tokens on `Specialized Knowledge Tasks'. Results suggest that combining rephrasing styles does not yield performance benefit on zero-shot tasks compared to just Q/A style.}
    \label{tab:specialized_knowledge_combination}
\end{table*}

\begin{table*}[t]
    \centering
    \scalebox{0.8}{
    \begin{tabular}{lcccccccc>{\columncolor{LightCyan}}c}
        \toprule
        Dataset (Real Tok.) & ARC-E & BoolQ & Wino. & PIQA & HellaSwag & TruthfulQA & OBQA & LogiQA & Avg \\
        \midrule
        Med+C4-35B &  59.8  &  57.0  &  55.7  &  74.6  &  44.5  &  36.5  &  23.8  &  21.5  &  46.7\\
        QA+C4-35B &  62.2  &  63.3  &  55.7  &  74.8  &  44.6  &  41.4  &  22.4  &  23.2  &  48.4\\
        Combined-1:1-35B &  60.6  &  60.2  &  57.7  &  73.8  &  43.7  &  40.2  &  22.0  &  22.1  &  47.5\\
        Combined-1:2-35B &  61.4  &  62.0  &  57.0  &  74.8  &  44.6  &  39.5  &  23.0  &  21.3  &  48.0 \\
        \bottomrule
    \end{tabular}}
    \caption{\textbf{Combining multiple styles:} Evaluation of $\sim1.3$B parameter LLMs trained for 150B tokens on General Understanding Tasks. 
    Results suggest that combining rephrasing styles does not yield performance benefit on zero-shot tasks compared to just Q/A style.}
    \label{tab:general_understanding_combination}
\end{table*}

\begin{figure}[t]
\centering
\includegraphics[width=\linewidth]{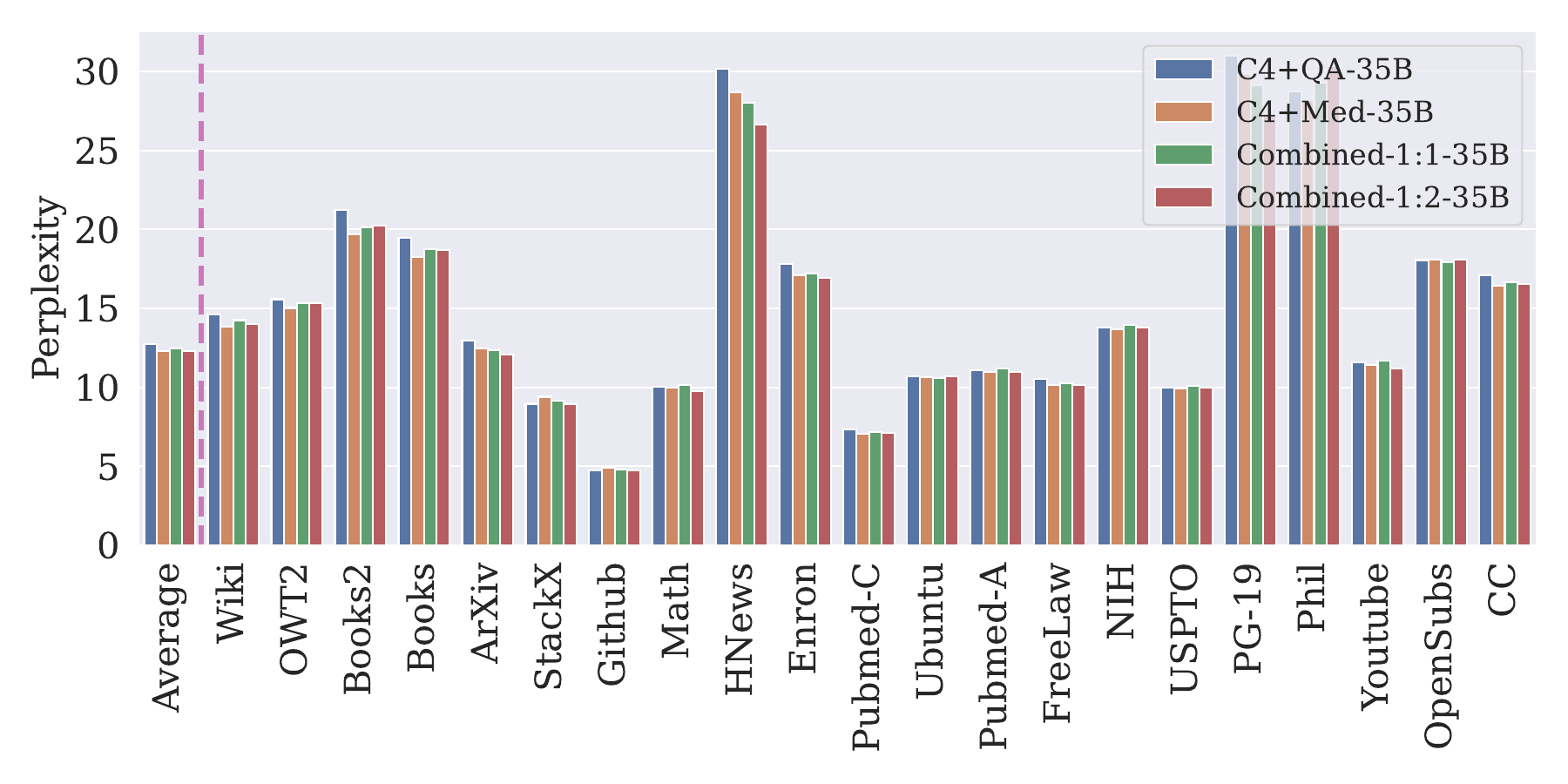}
\caption{\textbf{Combining multiple styles:} Perplexity across all domains of the Pile comparing combining multiple styles of synthetic data.  Models are 1.3B parameters trained for a total of 150B tokens. We see small perplexity improvements from combining multiple styles.}
\label{fig:combined_styles}
\end{figure}
Our findings in Tables~\ref{tab:general_understanding}--\ref{tab:specialized_knowledge}indicate that synthetic data using the QA prompt are sufficient for strong performance on QA tasks.  However, when evaluated on Pile perplexity, we observe significant degradation in perplexity across many sub-domains in Figure~\ref{fig:syn_comp}.  This is likely because synthetic data is very clean containing few special characters and being highly structured.  In contrast several sub-domains of the Pile such as OWT, and Hackernews have such special tokens.  On domains such as Philpapers and Gutenberg, we observe that dropping real C4 text from the pre-training data, and training on synthetic documents alone drops performance significantly (increase in perplexity). This is once again attributed to the fact that synthetic data does not contain certain `tags' and `styles' that are prevalent in real data scrapes, and emphasized how \ours~is a better strategy than pre-training on synthetic data alone.
In terms of performance on zero-shot tasks, we once again note that the presence of real data helps improve zero-shot performance in Tables~\ref{tab:general_understanding_150B_real_data},\ref{tab:specialized_knowledge_real_data}.
Since zero-shot tasks contain well-written Q/A pairs, this effect is not as evident as that for perplexities on real data.

\paragraph{\textbf{RQ2: Does a combination of multiple synthetic datasets improve performance?}}
We measure the impact of combining multiple synthetic styles with C4 for training.  We consider two variants: combining in a 1:1 ratio meaning that there are two copies of C4 to match two synthetic styles (medium and QA), and 1:2 ratio which combines only one instance of the C4 dataset.  For zero-shot QA tasks, our finding in Table~\ref{tab:specialized_knowledge_combination}-\ref{tab:general_understanding_combination}
 indicate lower performance than combining only QA and C4 data. Evaluations over the Pile are shown in Figure~\ref{fig:combined_styles}.  We notice that both the `Q/A' and `Wikipedia' paraphrases help improve performance on certain domains. For example, `Stackexchange', that has lots of question-answers benefits from the presence of synthetic data in Q/A style. Overall, we note that there is a small improvement on the average perplexity on the Pile by combining multiple styles.

\subsection{Method Ablations}

\paragraph{\textbf{RQ3: How important is to have a high-quality re-phraser?}}

\begin{figure}[t]
\centering
\includegraphics[width=\linewidth]{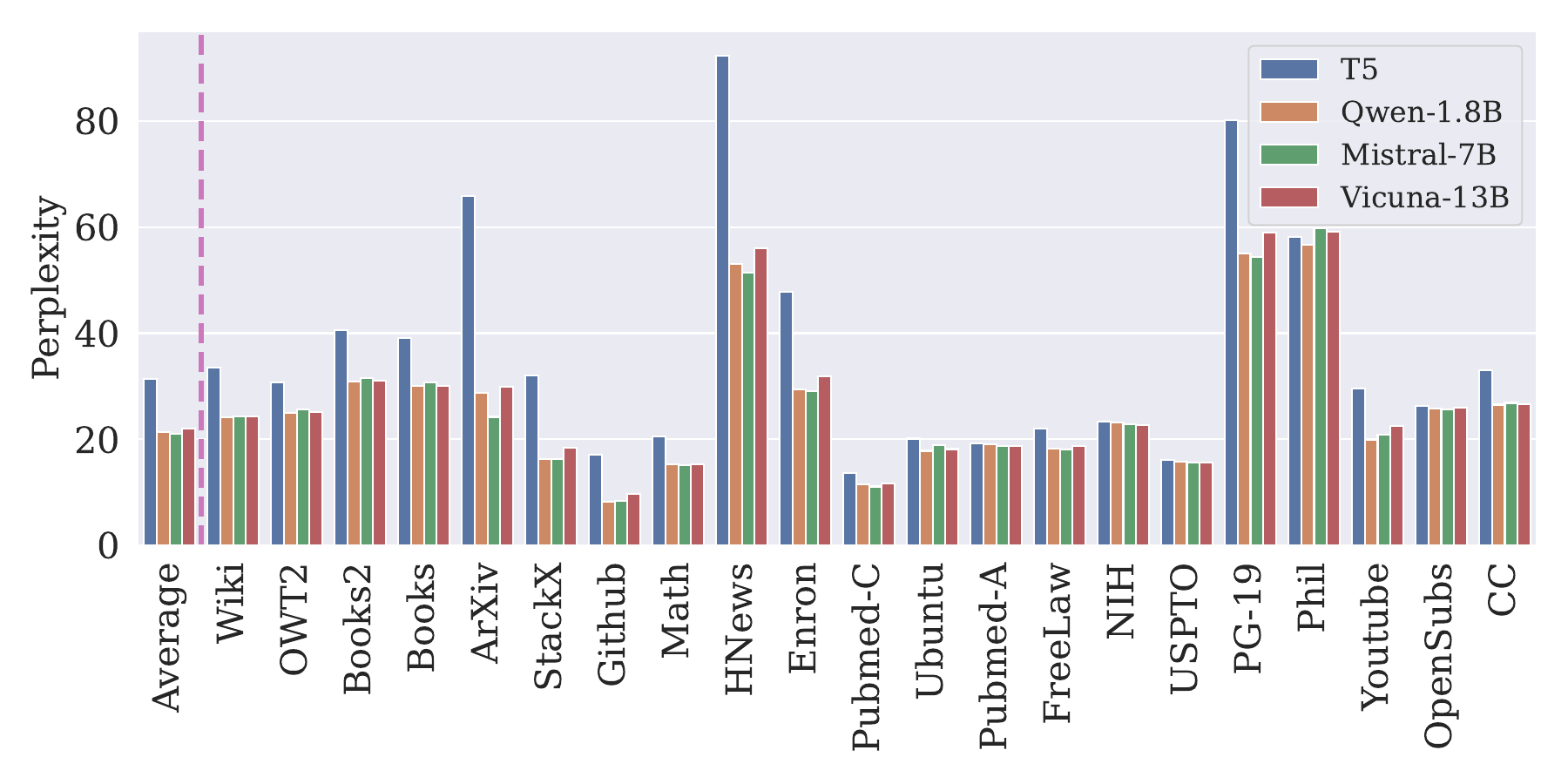}
\caption{\textbf{Importance of High Quality Paraphraser:} Perplexity across all the Pile domains for \ours~on data generated by different LLMs.
Results show that even small models like Qwen-1.8B can generate paraphrases of high quality. Though, a low quality rephraser like our fine-tuned T5-base model leads to significantly worse language modeling.
}
\label{fig:model_rephraser}
\end{figure}

\begin{figure}[t]
\centering
\includegraphics[width=\linewidth]{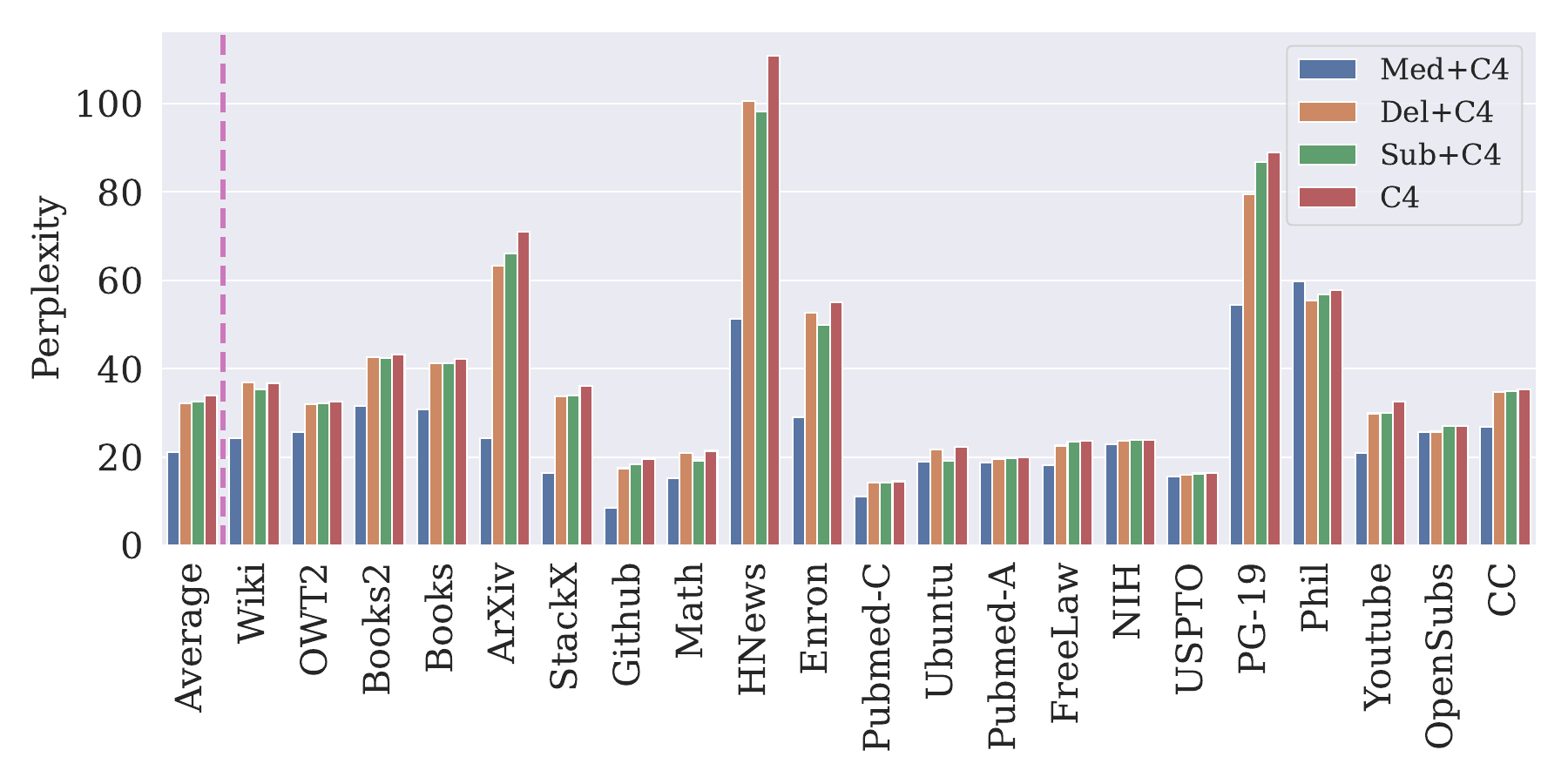}
\caption{\textbf{Is re-phrasing same as any augmentation?} We compare perplexity on the Pile for different augmemntation strategies.  350M parameter models are trained for a total of 15B tokens. \ours~(Medium + C4) performs significantly better than traditional augmentations.}
\label{fig:augs}
\end{figure}

To answer this,
we use data from four distinct re-phrasing models (T5-base~\citep{raffel2020exploring}, Qwen-1.8B-chat~\citep{qwen}, Mistral-7B-chat~\citep{jiang2023mistral}, and Vicuna-13B-chat-v1.3~\citep{vicuna2023}) and train a 345M model for 30B tokens.  We generate data from all models using the same prompt.  In case of the T5-base model, we finetune the model for 1 epoch on re-phrase pairs from the Vicuna-13b-chat  model. 
We find that pre-training on data generated by smaller re-phrase models like Qwen-1.8B and Mistral-7B achieve lower perplexity than 
Vicuna~13B
(Figure~\ref{fig:model_rephraser}).  
At the same time, our fine-tuned T5-base model performs significantly worse than the rest.
Even then, all rephrase models reduce perplexity over only real C4 data. 
It remains an open question to test the limits of how small can we train a paraphrase model that can generate high quality synthetic data to further scale the applicability of \ours.

\paragraph{\textbf{RQ4: Does synthetic data improve over augmentations?}}
Are the gains observed by pre-training on synthetic data the same as pre-training with augmentations? In order to test this, 
we consider two popular text augmentation baselines---synonym replacement and random deletion using the NL-Augmenter library~\citep{dhole2021nlaugmenter}.
We pre-train a 350M parameter model for 15B tokens in order to conduct this set of experiments. The total pool size is only about 1.5B tokens, meaning that the model would have to repeat data around 10 times during the pre-training phase, unless augmented over.
As seen in the perplexity analysis in Figure~\ref{fig:augs}, the models trained on augmented data perform significantly worse than those trained on combinations of real and synthetic data. This suggests that synthetic data enhances the learning process, and is not merely another form of augmentation.

\paragraph{\textbf{RQ5: How does the style of synthetic data impact performance on specialized domains?}}

 \begin{figure}[t]
\centering
\includegraphics[width=\linewidth]{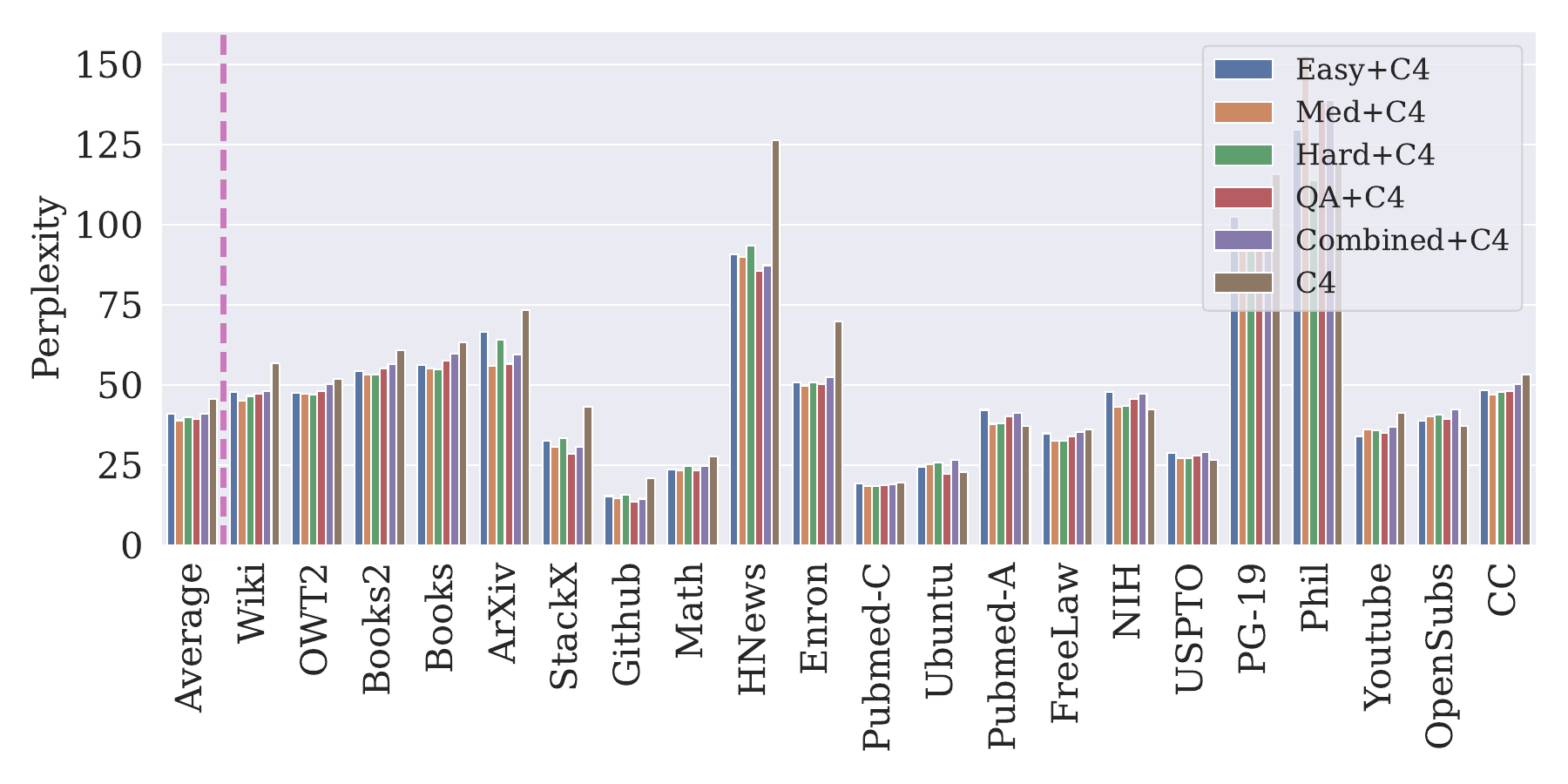}
\caption{\textbf{Impact of style of synthetic rephrases:} Perplexity across all domains of the Pile comparing different styles of synthetic data.  We train 128M parameter models for 3B tokens.}
\label{fig:styles}
\end{figure}
We compare the performance of various models trained on different styles of synthetic data.  In particular, we generate four styles of synthetic data (easy, medium, hard, and Q/A) and evaluate the performance of training on combinations of each style across Pile subsets. The prompts to generate these synthetic data styles are outlined in Appendix~\ref{app:prompt_styles}. Results corresponding to generations from a Vicuna-v1.3 model, and for a 128M model trained for 3B tokens are summarized in Figure~\ref{fig:styles}.
We see  that training with combinations of real C4 and synthetic data matching the style of the domain at evaluation improves performance.  However, we find that no single synthetic data style performs the best across all domains, resulting in similar performance across training with combinations of real C4 data and each synthetic style variant.  While knowing the best synthetic style to pre-train an LLM is impractical, an oracle that selects the best synthetic style across all domains will improve perplexity by 16\%---indicating the importance of training with diverse data styles for LLM generalization, even when the underlying knowledge stays the same.

\paragraph{\textbf{RQ6: Is there data leakage from the rephrase model to the trained model?}}

We investigate whether our synthetic data maintains similar semantic meaning while being stylistically different from the original C4 data and matching the style of different PILE domains.  We start by comparing pairs of examples of synthetic and real data to confirm the performance gain is not attributed to knowledge leakage from the rephrase models.  We take a subset of the first 1000 samples from each of the datasets. 

\begin{figure*}
  \centering
  \begin{subfigure}{0.45\linewidth}
    \includegraphics[width=\columnwidth]{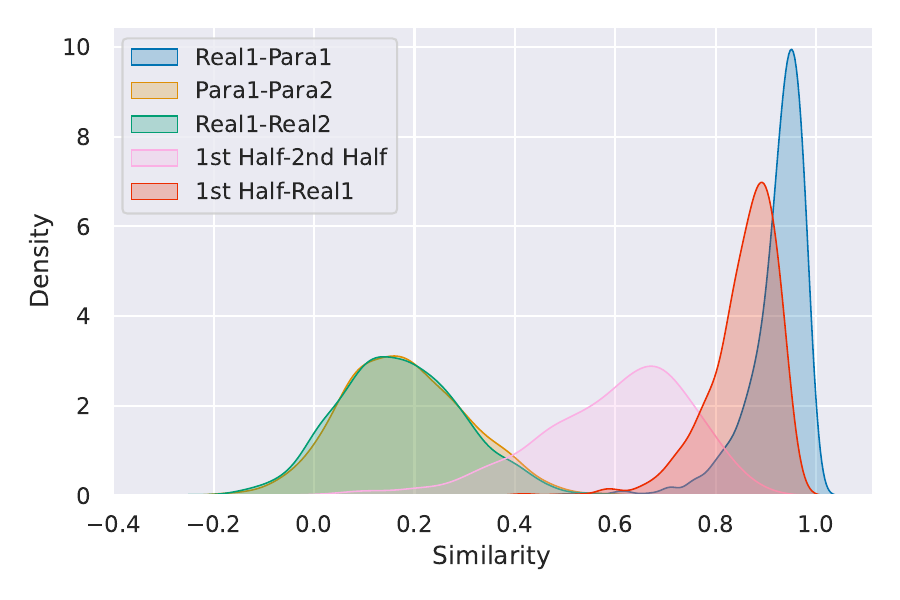}
    \caption{Cosine similarity Medium synthetic data}
  \end{subfigure}
  \hfill
  \begin{subfigure}{0.45\linewidth}
    \includegraphics[width=\columnwidth]{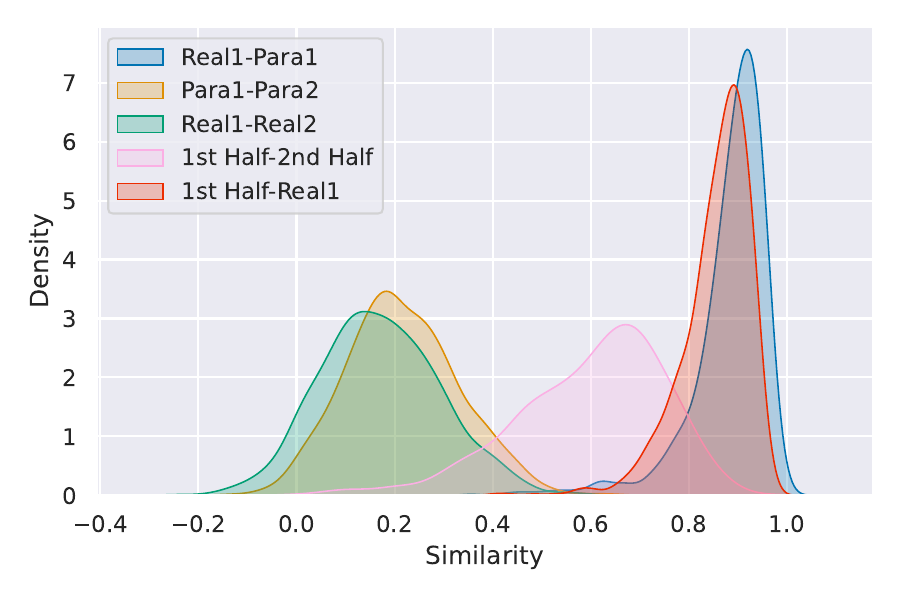}
    \caption{Cosine similarity QA synthetic data}
  \end{subfigure}
  \hfill
  \caption{Comparison between synthetic and real data from the C4 corpus showing that synthetic data maintains semantic meaning compared with the real C4 data and primarily changes style for (a) medium rephrases of C4, and (b) QA rephrases of C4.} %
  \label{fig:cosine_sim_style}
\end{figure*}

We show the cosine similarity of the sentence embeddings from a pre-trained BERT model trained with SimCSE objective \citep{gao2021simcse} for medium and qa prompts in  Figure~\ref{fig:cosine_sim_style}(a)  and (b).  When computing similarity, we remove outliers. Figures with distributions use a gaussian Kernel Density Estimator (KDE) to construct distributions for statistics from 1000 values. The cosine similarity of real-synthetic pairs is higher than several baselines including  two random real samples from C4, a continuation baseline which computes cosine between the first half of a sample and the full sample, and cosine similarity between the first half and second half of the same sample. High similarity indicates that the re-phrases maintain similar meaning to their real counterparts without adding information.

\section{Limitations and Opportunities}

\subsection{Cost Analysis}
\emph{Should you generate synthetic data, or just train longer on real data?}

The applications of \ours~lies in both paradigms---(i) low-resourced data settings such as a language model for Finnish language~\citep{luukkonen2023fingpt}, and (ii) data-rich settings such as training on the common crawl.
In the former, there is no alternative option of naively gathering more data, and hence, synthetic data is a natural solution that should outperform training on in-domain data alone. However, there is a significant interest in training language models on English, or more broadly, general web data. Is using synthetic data a viable option even in this paradigm?

Before, we dive into the feasibility of pre-training on synthetic data, we should acknowledge the results of Table~\ref{tab:general_understanding}. The TinyLlama model trained for 3 Trillion tokens also underperforms a model jointly trained on real and synthetic data. In fact, it performs quite comparably to the models that were trained for 300B tokens on just real data as well. This suggests that the ceiling for improvement by training for longer may not be that high (for a model of size 350M/1.3B parameters; larger models may benefit from training for longer).

To analyze this cost trade-off, we compare the cost of generating synthetic data, versus that of training a language model on extra data.
For our synthetic data generation experiments, we use the vLLM~\citep{kwon2023efficient} library for fast generation. In particular, we are able to generate 3M tokens per hour on a single A100 when using the Mistral-7B. Generating 85B tokens (as in our work) accounts for about 25K GPU hours. 

In comparison, on 64 A100s, we achieve a throughput of 0.5M tokens per second. Assuming training for 300B tokens, would mean 256 GPU days, accounting for about 6k GPU hours to train a single model. On the contrary, training a 13B model would take about 30K GPU hours. At the scale of training a 13B model, reducing the training cost by 3-10x can incorporate the cost overhead of training with synthetic data in a single run.

While the cost of generating high quality data is still relatively high, two important sources of improvement impact this cost analysis.  First, if we use the Qwen-1.8B model \cite{bai2023qwen} for rephrasing, we are able to get a 3x higher token throughput. As seen in our preliminary results in Fig~\ref{fig:model_rephraser}, the model pre-trained on rephrases generated by Qwen model performs comparably to that by the Mistral model. This reduces the cost of generation by 3x. More recent work in speculative decoding~\citep{liu2023online} and optimized inference~\citep{xia2024unlocking} suggest that we can leverage another 3-5x improvement in the generation cost. Hence, indeed, even at the scale of just 1.3B parameter model training, we can already improve upon the cost of pre-training using just real data.

Two additional important advantages of synthetic data generation that could not be accounted for in the discussion above:
\begin{enumerate}
    \item The cost of synthetic data generation is a one-time investment, and we may train many models of varying scales once the data is generated.
    \item Data generation is 100\% parallelizable, whereas training requires the availability of a big cluster with fast inter-node connectivity. This is much more expensive. On the other hand, generation can be thought of as a side process that can fill in the empty GPUs in any large-scale compute cluster, and runs on single GPU machines.
\end{enumerate}

\subsection{Diversity of Synthetic Generations}
Another limitation is enforcing the diversity in the generated data. This diversity comes from both the ``style'' and the ``knowledge'' contained in the generated data. Recent works~\citep{li2023self,li2023textbooks} used a selection of topics, or scenarios to seed the model to generate novel texts.  Still, a recent study by~\citet{padmakumar-etal-2023-investigating} showed that using language models for AI-assisted writing tends to reduce content diversity, particularly when using instruction-tuned models. While we used the paradigm of rephrasing specifically to mitigate the issues pertaining to the diversity of novel content generation, it remains for future work to assess the presence (or lack of) and impact of content diversity in paraphrase models.

\section{Conclusion}
Strong language models are being pre-trained on combinations of real and synthetic data. 
Using synthetic data enables baking in desirable attributes such as fairness, bias, and style (like instruction following) directly into the data, eliminating the need to adjust the training algorithm specifically. This offers an alternative approach to aligning language models to human values. 
The recent uptick in interest around synthetic data, especially for instruction-tuning language models, is noteworthy, with concurrent researchers also leveraging it for pre-training. 
As we transition into this paradigm, understanding the properties of the data fed to our models is paramount. 
This paper aims to be a comprehensive guide on employing different synthetic style data in LLM pre-training. We delve into its significance from two vantage points: (1) In scenarios with scarce high-quality data, synthetic rephrases offer more value than mere repetition of existing data; (2) Synthetic data can be a boon for generalization on different text domains, and for generating text in styles that are underrepresented in the pre-training dataset. 
As practitioners generate synthetic data for training models, they will be faced with important and expensive design choices---(i) How important is the quality of the synthetic data generator?; (ii) How to balance real and synthetic data? (iii) When does training on synthetic data reach a point of diminishing returns in terms of epochs? This work takes a first step towards answering these questions.

Conversely, it's essential to note the inherent limitations, and opportunities with synthetic data. We highlight two limitations: (1) cost of generation is still large and requires strong LMs, and (2) enforcing the diversity in the generated data is challenging.  In this work, we leverage the natural diversity of the web to generate synthetic ``re-phrases''. This limits the model from learning new ``knowledge'' and enhances the learning process only via the provision of high-quality inputs. Whereas past work required a more intricate understanding of the blind spots of the model, potentially biasing the knowledge contained in the pre-training data distribution. Nonetheless, we demonstrate the potential of synthetic data to improve LLM training efficiency both in compute and data size.

\bibliography{anthology,paper}
\bibliographystyle{colm2024_conference}

\newpage
\appendix
\section{Dataset Details}
\subsection{Training Dataset}

The primary pretraining corpus in our experiments is Colossal Clean Crawled Corpus (C4), a curated English text dataset comprising over 170 billion tokens. This corpus is derived from CommonCrawl, a common practice in the pretraining of LLMs \cite{brown2020language,raffel2020exploring,touvron2023llama}. This data source is also prominently featured in openly available LLM pretraining corpora, including The Pile \cite{gao2020pile} and RedPajama \cite{together2023redpajama}.
 There are different versions of CommonCrawl data and our selection of C4 for pretraining is driven by driven by its size and quality.

 We also compare with pre-training on the Refined Web corpus \cite{penedo2023refinedweb}.  The dataset is also derived from the CommonCrawl, however has a more stringent filtering process. Our selection of Refined Web is for comparing synthetic rephrases to high quality subsets of web data, which were shown to achieve similar performance compared with curated datasets \cite{penedo2023refinedweb}.  For our experiments we used the first $3050$ files and train for 300B tokens to match training on C4.  We aso conduct experiments with the first 1650 files to account for multiple epochs on the Refined Web dataset.  

\subsection{Pile Perplexity Evaluation}
For the evaluation phase, we employed 20 subsets from the Pile corpus. We excluded the Europarl subset because it contained non-English language. The subsets used are:
CC, StackExchange, Wikipedia, GitHub, PubMed Abstracts, Openwebtext2, Freelaw, Math, NIH, USPTO, Hackernews, Enron, Books3, PubMed Central, Gutenberg, Arxiv, Bookcorpus2, Opensubtitles, Youtubesubtitles, Ubuntu, and Philpapers. We take the first 10000 samples from each subset and split into documents of maximum length 1024.  The reported average in all perplexity plots is a weighted average over the perplexity of all domains according to the ratios in Table~\ref{tab:ratios}.  

\subsubsection{Pile Weighted Average Ratios}

We report the ratios for samples according to the first 10,000 documents from our Pile validation set in Table~\ref{tab:ratios}.  Note that there are some slight variations in the ratios compared with those reported in \citep{gao2020pile}, but most ratios are similar.
\begin{table}
    \centering
    \begin{tabular}{lcc}
    \toprule
       Dataset  & Validation Ratio (\%) & Published Ratio (\%) \\\hline
       ArXiv  & 10.4 & 9.0\\
       BookCorpus2 & 0.8 & 0.8\\
       Books3 & 11.8 & 12.1\\
       Pile-CC & 14.0 & 18.11\\
       Enron & 0.1 & 0.1\\
       EuroParl & 1.1 & 0.7\\
       FreeLaw & 5.3 & 6.1\\
       Github & 10.9 & 7.6\\
       Gutenberg & 1.5 & 2.2\\
       Hackernews & 0.6 & 0.6\\
       Dm Mathematics & 2.0 & 1.2\\
       NIH & 0.2 & 0.3\\
       OpenSubtitles & 1.3 & 1.6\\
       OpenWebText2 & 8.2 & 10.0\\
       PhilPapers & 0.7 & 0.4\\
       PubMed Abstracts & 0.7 & 3.1\\
       PubMed Central & 14.9 & 14.4\\
       StackExchange & 5.8 & 5.1\\
       Ubuntu & 1.3 & 0.9\\
       USPTO & 2.7 & 3.7\\
       Wikipedia & 3.4 & 1.5\\
       YoutubeSubtitles & 0.6 & 0.6\\
       \bottomrule
    \end{tabular}
    \caption{Pile ratios for our evaluation compared with published ratios}
    \label{tab:ratios}
\end{table}

\subsection{Zero-shot Evaluation Dataset}
We evaluate our models on a total of 13 different zero-shot benchmarks to assess their abilities across various natural language tasks. These benchmarks are categorized into two subsets: Specialized Knowledge and General Understanding.

\paragraph{Specialized Knowledge}
This subset comprises datasets that focus on domain-specific knowledge and expertise.

\begin{itemize}
\item \textbf{ARC Challenge (ARC-C)}: This dataset is part of the AI2 Reasoning Challenge (ARC)~\citep{allenai:arc}, containing science exam questions from grades 3 to 9. The ARC Challenge set includes more difficult questions that necessitate higher-order reasoning.
\item \textbf{SciQ}: A dataset of science exam questions, specifically designed to evaluate the ability of NLP models in understanding and reasoning within the scientific domain~\citep{SciQ}.
\item \textbf{PubMedQA}: This dataset focuses on biomedical literature and is designed to evaluate the understanding of medical and healthcare-related information~\citep{jin2019pubmedqa}.
\item \textbf{MathQA}: This dataset challenges models in mathematical problem-solving, requiring both numerical understanding and reasoning skills~\citep{amini-etal-2019-mathqa}.
\item \textbf{MMLU}: Multi-domain question answering, MMLU assesses the model's expertise over a wide range of specialized subjects, from professional domains to academia~\citep{hendryckstest2021}.
\end{itemize}

\paragraph{General Understanding}
This subset contains datasets that test general cognitive skills, language understanding, and common sense reasoning.

\begin{itemize}
\item \textbf{ARC Easy (ARC-E)}: The Easy set of the AI2 Reasoning Challenge~\citep{allenai:arc} features questions from the same source as ARC-C but are considered less challenging and do not require as advanced reasoning skills.
\item \textbf{BoolQ}: A dataset consisting of boolean (yes/no) questions, focusing on reading comprehension and general understanding of natural language text~\citep{clark-etal-2019-boolq}.
\item \textbf{Winogrande (Wino.)}: This dataset challenges models on common sense reasoning in a language context, focusing on pronoun disambiguation tasks~\citep{ai2:winogrande}.
\item \textbf{PIQA}: Physical Interaction Question Answering tests the understanding of everyday physical processes, an aspect of practical common sense~\citep{Bisk2020}.
\item \textbf{HellaSwag}: This dataset evaluates a model's ability to complete scenarios in a contextually and logically coherent manner, requiring both language understanding and common sense reasoning~\citep{zellers2019hellaswag}.
\item \textbf{TruthfulQA}: Focused on the generation of truthful, accurate answers, this dataset challenges models on their ability to discern and reproduce factually correct information~\citep{lin2021truthfulqa}.
\item \textbf{OpenBookQA (OBQA)}: OpenBookQA requires understanding a wide array of facts and concepts, thereby evaluating the model's broader knowledge and reasoning skills~\citep{OpenBookQA2018}.
\item \textbf{LogiQA-2}: This dataset involves logical reasoning, testing the model's capability to understand and apply logical constructs and principles~\citep{logiqa2}.
\end{itemize}

Each dataset in these subsets is carefully selected to challenge and evaluate specific aspects of natural language processing models, ranging from domain-specific knowledge in science, medicine, and mathematics, to broader skills like common sense reasoning and general language understanding.

\section{Filtering Details for Synthetic Data}
When generating synthetic paraphrases using language models, we occasionally encounter the challenge of extraneous introductions in the generated outputs. Such paraphrases might commence with phrases like "Here's a paraphrase...", "The following..." or even contain keywords such as "high-quality English". To mitigate this, we've developed a method to filter and refine the synthetic outputs.

\subsection{Methodology}

The primary function, \texttt{remove\_unwanted\_part}, starts by splitting the input data into individual sentences. If the first sentence contains delimiters such as "\textbackslash n\textbackslash n" (indicating a new paragraph) or ":", the function checks the segment preceding the delimiter for the aforementioned unwanted elements. If these elements are detected, the preceding segment is removed. The entire revised content is then reconstructed and returned. In cases where no modifications are applicable, but we still have the flagged keywords, we remove the paraphrase completely. To achieve this: 

\begin{enumerate}
    \item Split the input data into individual sentences using the NLTK's sentence splitter function.
    \item Examine the first sentence for the presence of delimiters.
    \item If a delimiter is detected, check the preceding segment for unwanted elements.
    \item If unwanted elements are found, discard the preceding segment (before an occurrence of \texttt{"\textbackslash n\textbackslash n"} or \texttt{":"}).
    \item Modify and return the filtered paragraph.
\end{enumerate}

Based on manual inspection we found that the error rate (occurrence of sentences with unwanted elements) after the modification is less than 0.1\%.

\section{Properties of Synthetic Corpus}
\label{sec:synth_properties}
To understand the properties of synthetic data generated from the rephrase model that lead to better pre-training performance, we compare the semantic similarity, syntactic complexity, and diversity between synthetic data, C4 data, and data from the Pile.  Our primary focus is answering the following questions about synthetic data: (i) Do models trained on synthetic data perform better due to information leakage from the rephrase model? (ii) Does the rephrase model accurately capture multiple styles? (iii) What attributes of synthetic data make it high quality? Our investigation helps address what data is beneficial for better generalization to specific domains, and quantify the importance of data variability and quality.

\subsection{Experimental Setup}
We take a subset of the first 1000 documents from each of the datasets.  For synthetic comparisons with real C4 data, we take pairs of samples, while for Pile subsets, we take the first 1000 samples from the test subset. When computing dataset quality statistics, we remove outliers more than two standard deviations in metric value.  When the number of samples from the Pile subset was fewer than 1000, we split samples. Figures with distributions use a Gaussian Kernel Density Estimator (KDE) to construct distributions for statistics from 1000 values.

\subsection{Semantic Properties}

\begin{figure*}
  \centering
 
    \includegraphics[width=4.5in]{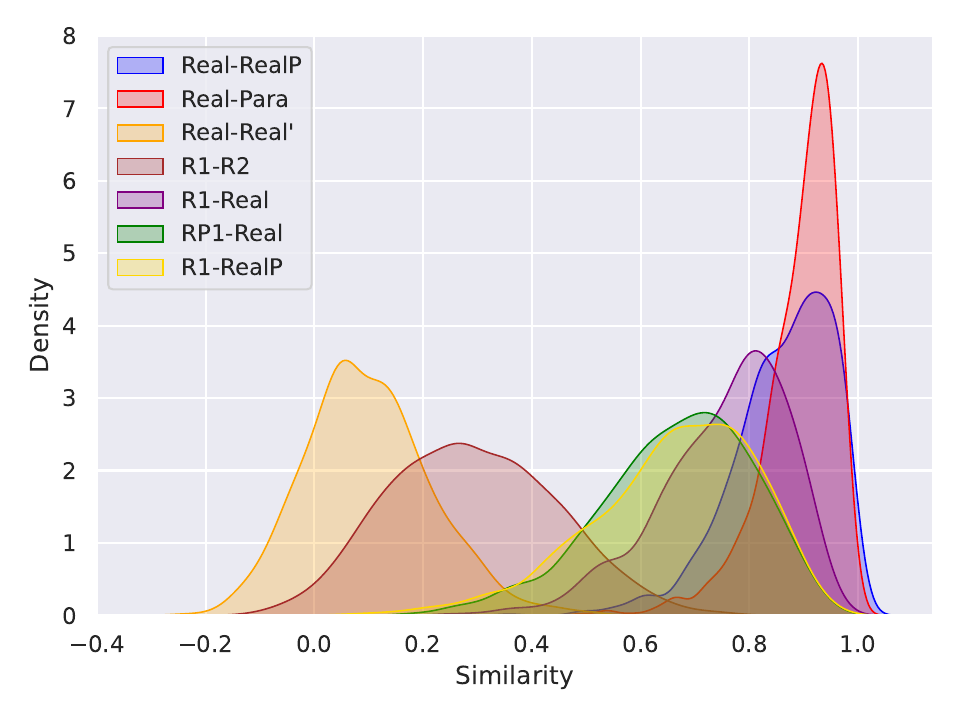}
    \caption{Cosine similarity medium synthetic MRPC rephrases}

  \label{fig:cosine_sim_mrpc_style}
\end{figure*}

In Section~\ref{sec:ablations}, we compared pairs of examples of synthetic and real data to confirm the performance gain is not attributed to knowledge leakage from the rephrase models using a pre-trained BERT model trained with SimCSE objective \citep{gao2021simcse} for medium and qa prompts in  Figure~\ref{fig:cosine_sim_style}(a)  and (b).  We additionally compare the similarity of synthetic rephrases and actual rephrases using the MRPC corpus in Figure~\ref{fig:cosine_sim_mrpc_style}(c).  We denote this additional comparison by RealP (real paraphrase), while maintaining comparison of splits of the sentence: R1 and R2.  Synthetic rephrases have similar cosine similarity on average and lower spread compared with true rephrases according in the MRPC corpus.

\begin{figure*}
  \centering
  \begin{subfigure}{0.45\linewidth}
    \includegraphics[width=\columnwidth]{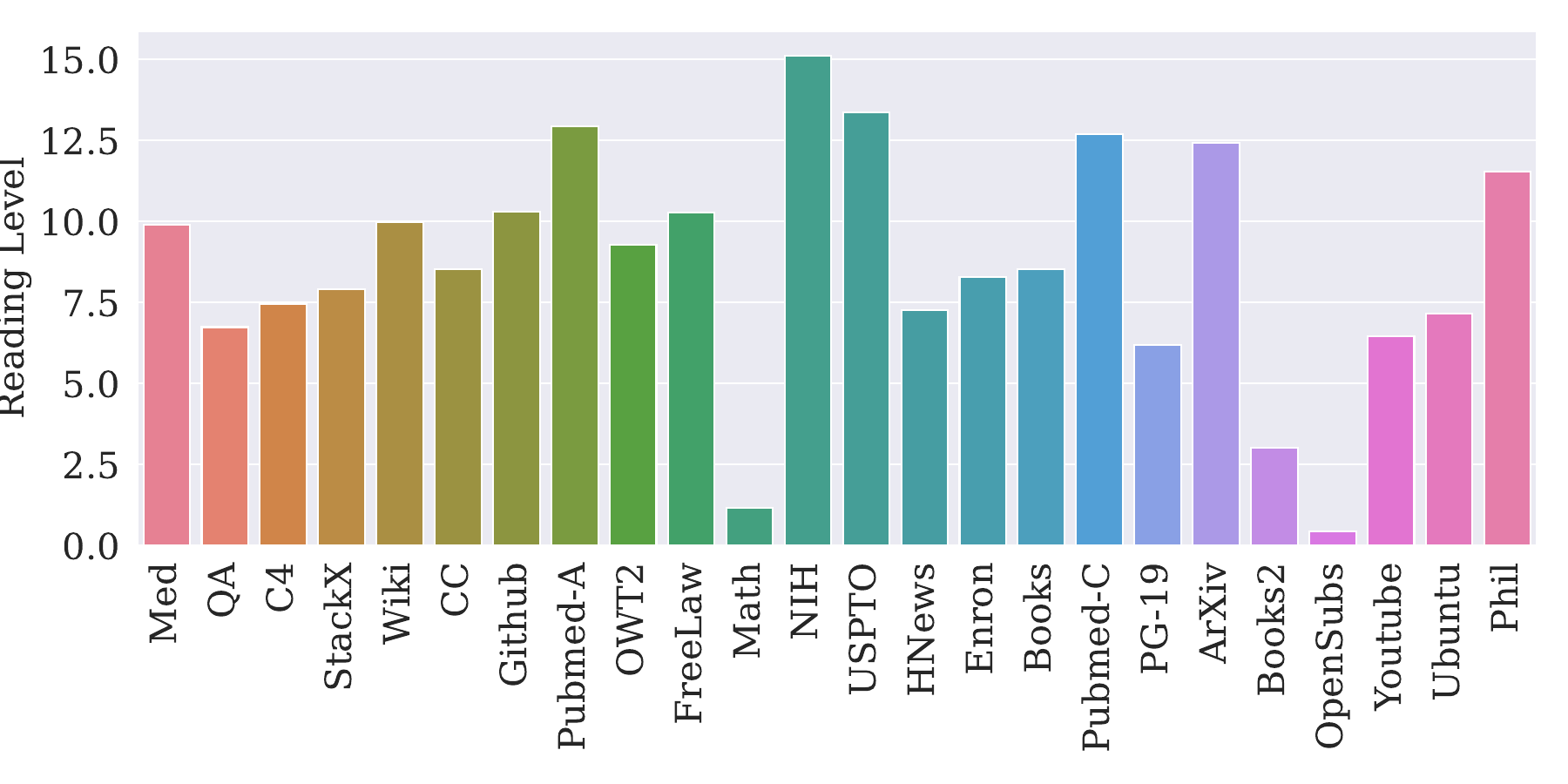}
    \caption{Flesch-Kincaid Reading Level}
  \end{subfigure}
  \hfill
  \begin{subfigure}{0.45\linewidth}
      \includegraphics[width=\columnwidth]{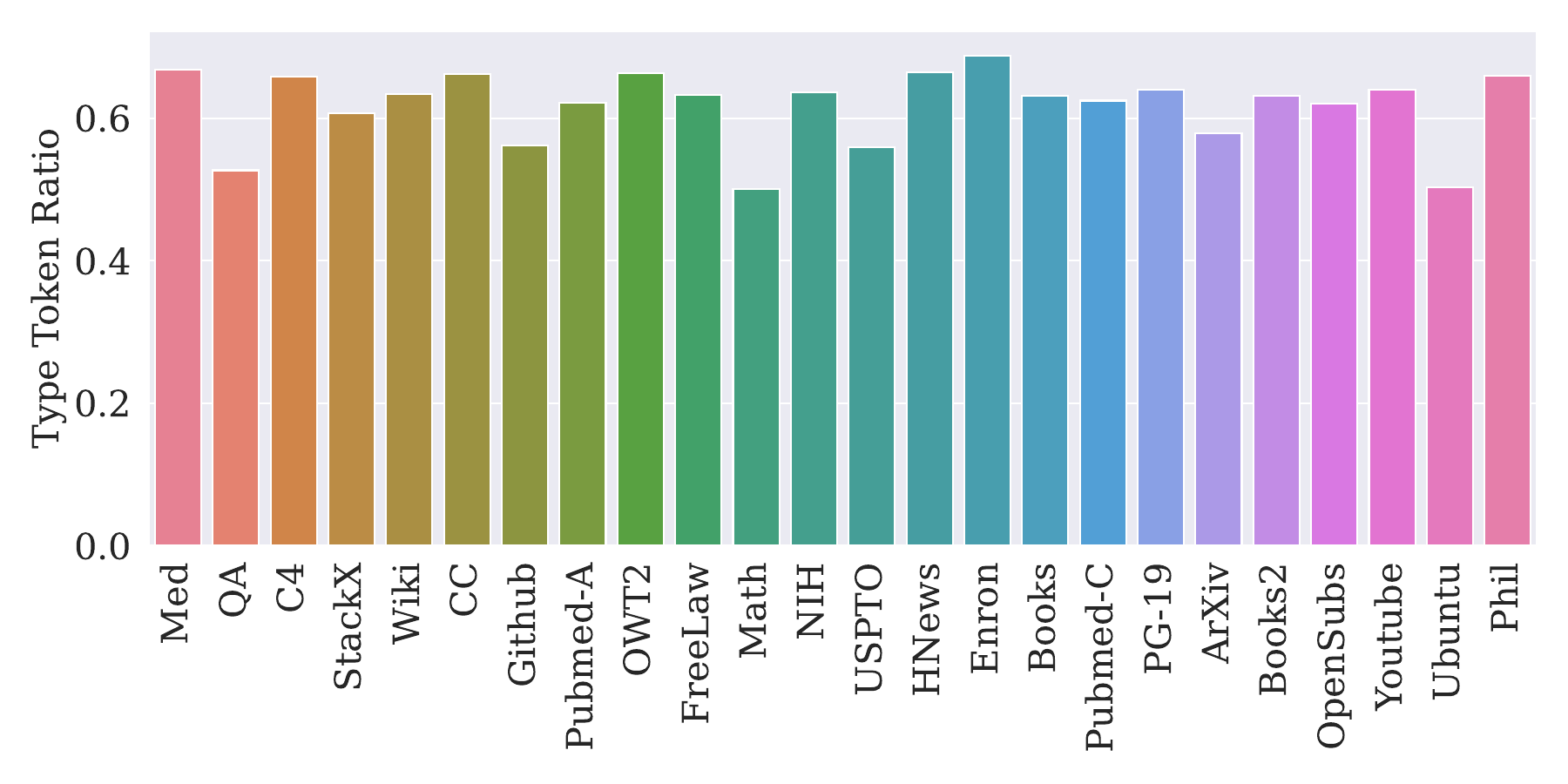}
        \caption{Type token ratio}
  \end{subfigure}
  \caption{Comparison of readability and diversity (ttr) of synthetic data compared with C4 and different subsets of the Pile.}
  \label{fig:readability}
\end{figure*}

As the semantic information is similar between C4 and our synthetic data, we further investigate stylistic differences in the data. Figure~\ref{fig:readability}(a) shows the Flesch–Kincaid reading levels for different rephrase styles, and the Pile.  Our findings indicate that C4 is on the low end of reading level (7-8).  In contrast,  medium increases the reading level to 10, and qa synthetic variants further reduces the reading level to 6.  Medium synthetic data matches the reading level of Wikipedia, and other high reading level datasets yielding better performance on these domains. On QA synthetic data, we observe reduced reading level. This is because we observed that sentences are typically split into question and answer leading to shorter setnences compared with in the original text and medium style rephrases.  This leads to lower metric values for many of the metrics. For type token ratio, we note that the diversity is quite similar between medium and most subsets of the Pile.  The QA dataset has particularly low TTR matching ubuntu, github, and math as these are more similar to QA format datasets and have heavy repetition of the Question, and Answer format.

\subsection{Syntactic Properties}
\begin{figure*}
  \centering
  \begin{subfigure}{0.45\linewidth}
    \includegraphics[width=\columnwidth]{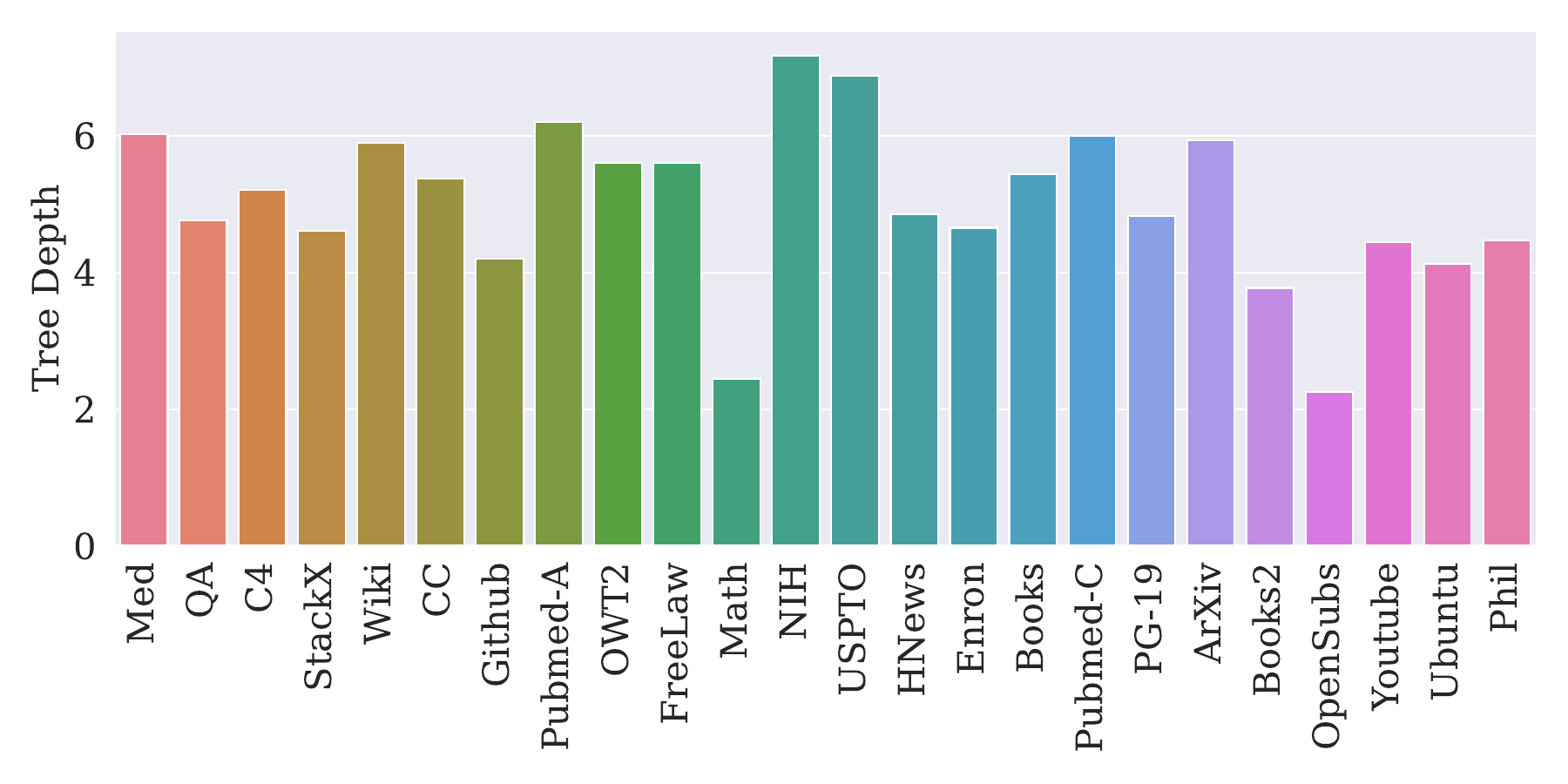}
    \caption{Tree Depth}
  \end{subfigure}
  \hfill
  \begin{subfigure}{0.45\linewidth}
    \includegraphics[width=\columnwidth]{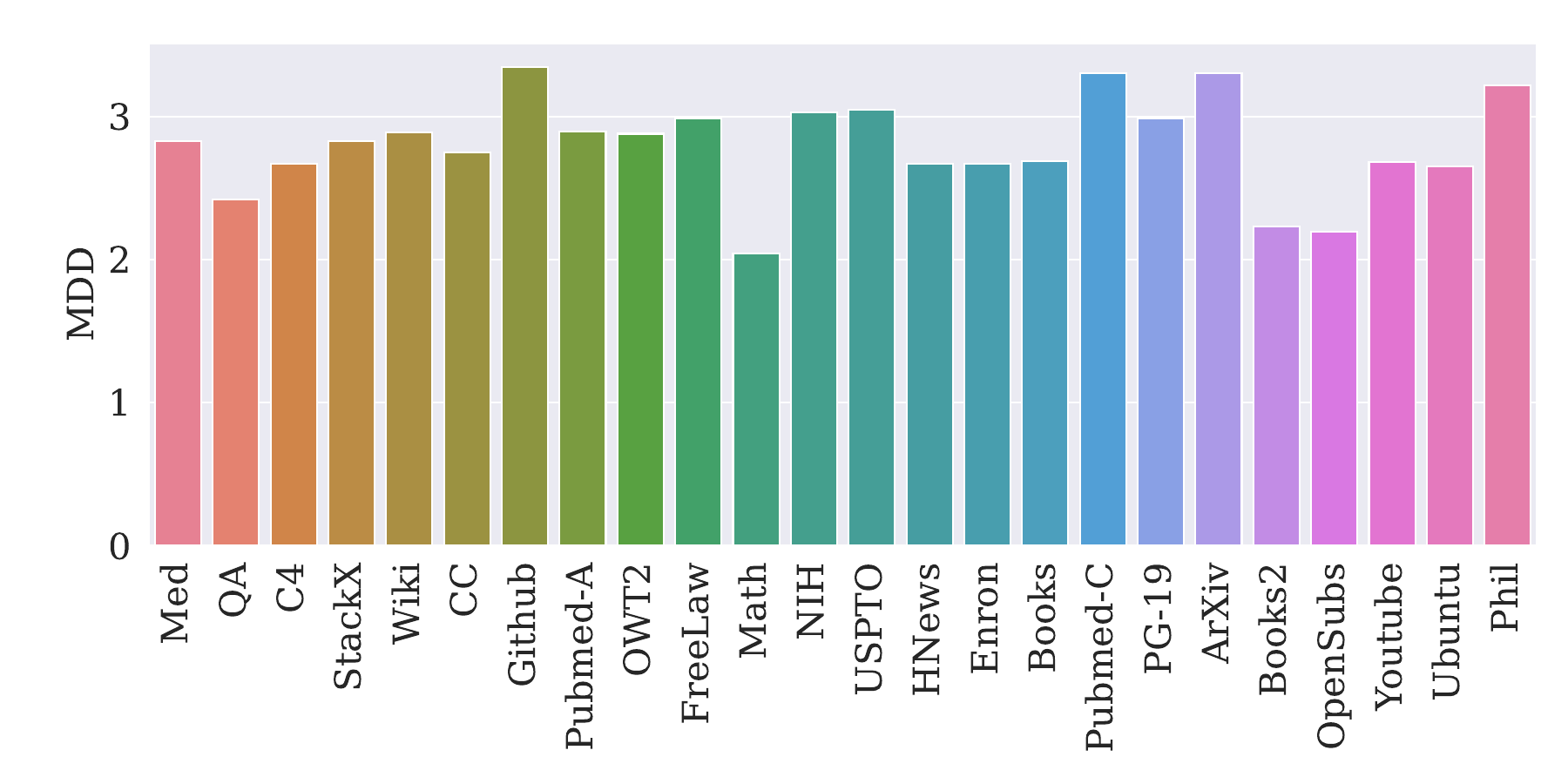}
    \caption{Mean Dependency Distance}
  \end{subfigure}  
  \caption{Comparison between synthetic and real data from the C4 corpus showing that synthetic data have higher syntactic complexity indicated by higher average tree depth, and higher mean dependency distance (MDD).}
  \label{fig:syntactic_complexity}
\end{figure*}

Finally, we compare the mean tree depth (measured by the mean over setences of the depth of the dependency tree),  and mean dependency distance (measured as the average dependency distance of any pair of words within a sentence) in Figure~\ref{fig:syntactic_complexity}, which have been shown to be good measures of syntactic difficulty \cite{futrell2015large,gibson2000dependency,oya2021three}.    We find similar trends as for reading level and TTR diversity where mediums tyle increase depth, mdd, and syntactic complexity in general.  We find again that QA style reduces this complexity.

\section{Evaluation Metrics}\label{app_eval}

The metric utilized for evaluation is the \textit{macro token level perplexity}. Given a batch of encoded texts, the perplexity at the token level was computed as follows:

Given the accumulated loss over the entire dataset, denoted as \( L \), and the total number of tokens, represented by \( T \), the macro token-level perplexity, denoted as \( \mathcal{P} \), is calculated as:

\begin{equation}
\mathcal{P} = \exp\left(\min\left(20, \frac{L}{T}\right)\right)
\end{equation}

Where:
\begin{itemize}
    \item \( \exp \) is the exponential function.
    \item \( L \) is the cumulative loss over all shifted logits and labels in the dataset.
    \item \( T \) is the total number of tokens in the dataset.
\end{itemize}

The value of 20 acts as an upper limit to stabilize the metric in cases of high loss values.

\section{Additional Results for Smaller Model and Token Sizes}

\subsection{Results for 350M Models Trained for 75B Tokens}

 \begin{figure}[t]
    \centering
    \includegraphics[width=\linewidth]{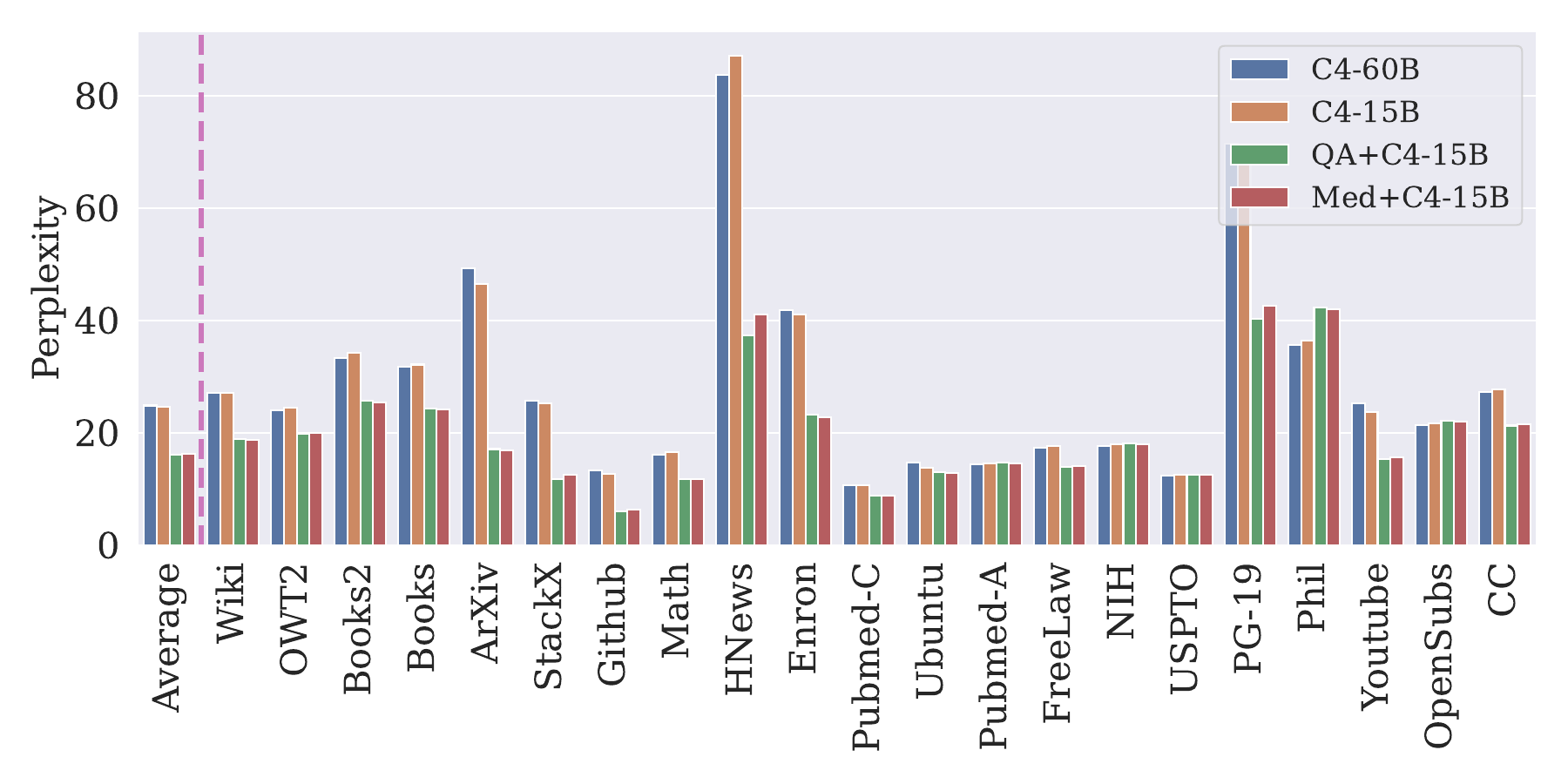}
    \caption{Perplexity across all domains of the Pile comparing combining multiple styles of synthetic data.  Models are 350M parameters trained for a total of 75B tokens.}
    \label{fig:350M_75K}
\end{figure}

\begin{table*}[t]
    \centering
    \scalebox{0.8}{
    \begin{tabular}{lccccc>{\columncolor{LightCyan}}c}
        \toprule
        Dataset (Real Tok.) & ARC-C & SciQ & PubMedQA & MathQA & MMLU & Avg \\
        \midrule
        C4-15B & 21.2  &  77.1  &  50.6  &  22.2  &  23.1  &  38.8\\
        C4-60B & 23.4  &  76.2  &  46.4  &  22.0  &  23.0  &  38.2\\
        QA+C4-15B &  24.4  &  79.8  &  56.0  &  21.7  &  22.9  &  41.0\\
        Med+C4-15B & 22.7  &  74.5  &  53.6  &  22.0  &  23.1  &  39.2\\
        \bottomrule
    \end{tabular}}
    \caption{Evaluation of 350M parameter LLMs trained for 75B tokens on Specialized Knowledge Tasks. This table presents the performance on tasks that require specific domain knowledge such as science, medicine, mathematics, and logic.}
    \label{tab:specialized_knowledge_350M}
\end{table*}

\begin{table*}[t]
    \centering
    \scalebox{0.8}{
    \begin{tabular}{lcccccccc>{\columncolor{LightCyan}}c}
        \toprule
        Dataset (Real Tok.) & ARC-E & BoolQ & Wino. & PIQA & HellaSwag & TruthfulQA & OBQA & LogiQA & Avg \\
        \midrule
        C4-18B &  50.5  &  52.8  &  53.0  &  69.8  &  35.6  &  37.8  &  18.6  &  23.0  &  42.6\\
        C4-75B &  51.4  &  53.4  &  51.6  &  70.3  &  36.1  &  39.0  &  17.4  &  22.6  &  42.7\\
        QA+C4-18B & 53.4  &  60.7  &  52.2  &  70.0  &  36.3  &  40.0  &  17.6  &  22.3  &  44.1\\
        Med+C4-18B & 50.6  &  57.3  &  53.6  &  70.8  &  36.1  &  36.9  &  18.6  &  22.0  &  43.2\\
        \bottomrule
    \end{tabular}}
    \caption{Evaluation of 350M parameter LLMs trained for 75B tokens on General Understanding Tasks. This table shows the performance across various datasets, focusing on general reasoning, language understanding, and common sense comparing training .}
    \label{tab:general_understanding_350M}
\end{table*}

We train models at smaller scales and demonstrate improvement. In particular we train a 350M GPT-2-medium architecture for a total of 75B tokens.  We show Pile perplexity averaged across the 21 domains is much lower than for that of the model trained only on C4 in Figure~\ref{fig:350M_75K}, and even lower than 1.3B models trained only on C4 in Figure~\ref{fig:lower_right_image}.  We also show an increase of $1.5\%$  across general understanding language tasks, and roughly $3\%$ on specialized knowledge tasks in Tables~\ref{tab:specialized_knowledge_350M}--\ref{tab:general_understanding_350M} when adding QA rephrases. We also experimented with medium rephrases at this smaller scale. Our findings indicate that the high quality provided by medium rephrases improves performance over only C4, however matching the style as indicated by QA rephrase performance further improves performance.

\subsection{Results for 1.3B Models Trained for 150B Tokens}

\begin{figure}[t]
\centering
\includegraphics[width=\linewidth]{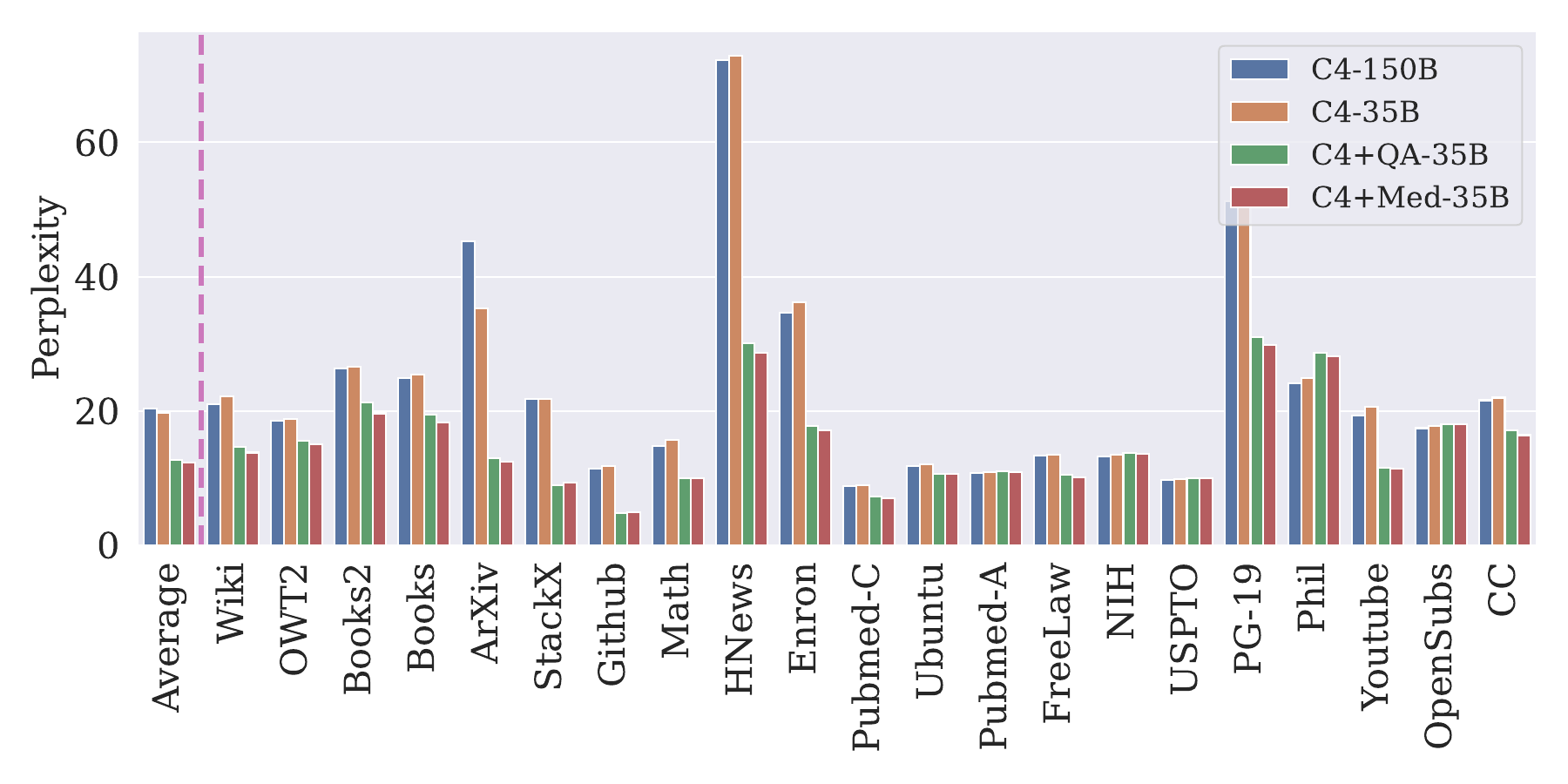}
\caption{Perplexity across all domains of the Pile comparing combining multiple styles of synthetic data.  Models are 350M parameters trained for a total of 75B tokens.}
\label{fig:all_perplexity_150k}
\end{figure}

We additionally train 1.3B GPT-2-XL models at 150B tokens, reducing the number of steps by half.  We show Pile perplexity averaged across the 20 domains is much lower than for that of the model trained only on C4 in Figure~\ref{fig:all_perplexity_150k}, and even lower than 1.3B models trained only on C4 in Figure~\ref{fig:lower_right_image} for twice as long.  We also show an increase of $2\%$  across specialized knowledge tasks, and roughly $2.5\%$ on general understanding tasks in Tables~\ref{tab:specialized_knowledge_150B}-\ref{tab:general_understanding_150B} when adding QA rephrases. We also experimented with medium rephrases at this smaller scale, and report similar findings consistent with other small-scale experiments.

\begin{table*}[t]
    \centering
    \scalebox{0.8}{
    \begin{tabular}{lccccc>{\columncolor{LightCyan}}c}
        \toprule
        Dataset (Real Tok.) & ARC-C & SciQ & PubMedQA & MathQA & MMLU & Avg \\
        \midrule
        C4-35B &  27.0  &  83.4  &  55.0  &  22.5  &  24.3  &  42.4\\
        C4-150B  &  25.9  &  83.8  &  55.4  &  23.5  &  25.4  &  42.8\\
       Med+C4-35B & 27.2  &  82.2  &  46.2  &  23.1  &  25.2  &  40.8\\
        QA+C4-35B & 29.0  &  85.1  &  62.2  &  22.5  &  26.1  &  45.0\\
        \bottomrule
    \end{tabular}}
    \caption{Evaluation of $\sim1.3$B parameter LLMs trained for 150B tokens on Specialized Knowledge Tasks. This table presents the performance on tasks that require specific domain knowledge such as science, medicine, mathematics, and logic.}
    \label{tab:specialized_knowledge_150B}
\end{table*}

\begin{table*}[t]
    \centering
    \scalebox{0.8}{
    \begin{tabular}{lcccccccc>{\columncolor{LightCyan}}c}
        \toprule
        Dataset (Real Tok.) & ARC-E & BoolQ & Wino. & PIQA & HellaSwag & TruthfulQA & OBQA & LogiQA & Avg \\
        \midrule
        C4-35B &  58.6  &  55.2  &  56.1  &  73.9  &  44.5  &  36.0  &  22.2  &  22.8  &  46.2\\
        C4-150B &  59.1  &  54.4  &  56.4  &  74.5  &  44.9  &  34.3  &  22.2  &  22.1  &  46.0\\
        Med+C4-35B &  59.8  &  57.0  &  55.7  &  74.6  &  44.5  &  36.5  &  23.8  &  21.5  &  46.7\\
        QA+C4-35B &  62.2  &  63.3  &  55.7  &  74.8  &  44.6  &  41.4  &  22.4  &  23.2  &  48.4\\
        \bottomrule
    \end{tabular}}
    \caption{Evaluation of $\sim1.3$B parameter LLMs trained for 150B tokens on General Understanding Tasks. This table shows the performance across various datasets, focusing on general reasoning, language understanding, and common sense comparing training .}
    \label{tab:general_understanding_150B}
\end{table*}

\section{LLM Leaderboard Few-shot Results}
In our main experiments in Section~\ref{sec:results} we demonstrate that LLMs trained with synthetic rephrases are a better backbone for  zero-shot question-answering tasks as the model learns the question-answer format and style during pre-training.  In this section, we show that improvements from pre-training on synthetic rephrases are still present even in few-shot settings where the model has access to test samples.  To study few-shot performance, we evaluate on six tasks present in the OpenLLMLeaderboard\footnote{\url{https://huggingface.co/spaces/HuggingFaceH4/open\_llm\_leaderboard}}:

\begin{enumerate}
    \item ARC-Challenge (25 shot)
    \item HellaSwag (10 shot)
    \item MMLU (5 shot)
    \item Truthful-QA (5 shot)
    \item Winogrande (5 shot)
    \item GSM8k (5 shot)
\end{enumerate}

We evaluate two models trained for 300B and 350B tokens corresponding to roughly 85B and 100B unique C4 tokens respectively.  Our findings show substantial improvements on the ARC-challenge benchmark, and Truthful-QA conssitent in the zero-shot settings and comparable performance across other datasets.  Our models also perform better than the publicly released Falcon-1.3B model trained on the Refined Web dataset, and the Pythia-1.4B model, which was trained on Pile.  

\begin{table*}
    \centering
    \resizebox{\textwidth}{!}{
    \begin{tabular}{lcccccc>{\columncolor{LightCyan}}c}
        \toprule 
     Dataset & ARC  &  Hellaswag  &  MMLU  &  TruthfulQA  &  WinoGrande  &  GSM8K  & Avg \\\midrule
     C4 &  31.7  &  62.1  &  26.7  &  33.4  &  57.9  &  0.9  & 35.5 \\
    Falcon-RW &   35.1  &  63.6  &  25.3  &  36.0  &  62.0  &  0.5  & 37.1 \\
     Pythia-1.4b-Pile &    32.7  &  55.0  & 25.6   & 38.7   &  57.3  &  0.8  & 35.0 \\
     \midrule
     QA+C4-85B (300K) &  36.4  &  60.9  & 25.5  &  40.6  &  59.4  &  0.4  & 37.2\\
     QA+C4-100B (350K) &   35.5 &  60.5  & 26.8 &  40.6  & 61.3   & 0.3   & 37.5 \\
     \bottomrule
    \end{tabular}
    }
    \caption{1.3B 300K LLM Leaderboard Eval. Evaluation is done on a single seed (1234).}
    \label{tab:llm_leaderboard}
\end{table*}

\clearpage

\section{Rephrase Prompt Templates}
\label{app:prompt_styles}
We detail the prompts given to the Mistral-7B model to generate synthetic versions of the C4 dataset in specific styles.  \emph{Note: there are slight variations in the prompt that were used for other frozen LLMs, and no prompt was used for the T5 model.}

\subsection*{Easy Style}
A style designed to generate content understandable by toddlers.
\begin{tcolorbox}[colback=gray!20, colframe=gray!75, rounded corners, sharp corners=northeast, sharp corners=southwest]
\texttt{A chat between a curious user and an artificial intelligence assistant. The assistant gives helpful, detailed, and polite answers to the questions. USER: For the following paragraph give me a paraphrase of the same using a very small vocabulary and extremely simple sentences that a toddler will understand:}
\end{tcolorbox}

\subsection*{Hard Style}
A style designed to generate content comprehensible primarily to scholars using arcane language.
\begin{tcolorbox}[colback=gray!20, colframe=gray!75, rounded corners, sharp corners=northeast, sharp corners=southwest]
\texttt{A chat between a curious user and an artificial intelligence assistant. The assistant gives helpful, detailed, and polite answers to the questions. USER: For the following paragraph give me a paraphrase of the same using very terse and abstruse language that only an erudite scholar will understand. Replace simple words and phrases with rare and complex ones:}
\end{tcolorbox}

\subsection*{Medium Style}
A style designed to generate content comparable to standard encyclopedic entries.
\begin{tcolorbox}[colback=gray!20, colframe=gray!75, rounded corners, sharp corners=northeast, sharp corners=southwest]
\texttt{A chat between a curious user and an artificial intelligence assistant. The assistant gives helpful, detailed, and polite answers to the questions. USER: For the following paragraph give me a diverse paraphrase of the same in high quality English language as in sentences on Wikipedia:}
\end{tcolorbox}

\subsection*{Q/A Style}
A style intended to convert narratives into a conversational format.
\begin{tcolorbox}[colback=gray!20, colframe=gray!75, rounded corners, sharp corners=northeast, sharp corners=southwest]
\texttt{A chat between a curious user and an artificial intelligence assistant. The assistant gives helpful, detailed, and polite answers to the questions. USER: Convert the following paragraph into a conversational format with multiple tags of "Question:" followed by "Answer:":}
\end{tcolorbox}

\clearpage

\section{Rephrase Examples}
\textbf{Samplesfrom the MRPC corpus generated by the Mistral-7B model.}

\subsection*{Original}
\begin{tcolorbox}[colback=gray!20, colframe=gray!75, rounded corners, sharp corners=northeast, sharp corners=southwest]
\texttt{The stock rose \$2.11, or about 11 percent, to close Friday at \$21.51 on the New York Stock Exchange.\\\\
    Revenue in the first quarter of the year dropped 15 percent from the same period a year earlier.}
\end{tcolorbox}

\subsection*{Medium Style}
\begin{tcolorbox}[colback=gray!20, colframe=gray!75, rounded corners, sharp corners=northeast, sharp corners=southwest]
\texttt{The stock experienced an increase of approximately 11 percent, closing at \$21.51 on the New York Stock Exchange on Friday, with a rise of \$2.11.\\\\
During the initial three months of the current year, there was a 15 percent decrease in revenue compared to the corresponding quarter of the previous year.}
\end{tcolorbox}

\subsection*{Q/A Style}
\begin{tcolorbox}[colback=gray!20, colframe=gray!75, rounded corners, sharp corners=northeast, sharp corners=southwest]
\texttt{Question: What was the stock's closing price on Friday? Answer: \$21.51 Question: How much did the stock rise on Friday? Answer: \$2.11 or about 11 percent.\\\\
Question: What was the revenue drop in the first quarter compared to the same period last year? Answer: The revenue dropped 15 percent.}
\end{tcolorbox}

\clearpage

\textbf{Samples from the C4 corpus generated by the Mistral-7B model.}

\subsection*{Original}
\begin{tcolorbox}[colback=gray!20, colframe=gray!75, rounded corners, sharp corners=northeast, sharp corners=southwest]
\texttt{First round on stress at work survey. Answering the questionnaire is voluntary and all answers will be saved anonymously. Please fill in this questionnaire only if you have some work experience, part-or full time. Otherwise, you will not be able to answer some of the questions! Here is a the link to all language version.\\\\
Not that there's a thing wrong with frozen burgers. The key here is the meat seasonings, which are pretty strong and spicy and just GOOD, something else I think is really necessary in a turkey burger because ground turkey otherwise can be kind of flavorless. You'll need ground turkey, onion powder, chili powder, salt, pepper, and cayenne pepper for the burgers. Then the mayo takes garlic and onion. Then we need buns, clearly, swiss cheese, lettuce, and onion. I LOVE tomatoes but sometimes find that they get in the way of other flavors, so I left them off of this burger. Add them if you'd like to your array of toppings! First, we'll make the mayo. Grate the garlic directly into the mayo, add a pinch of salt, and squeeze in the lemon juice. Stir. Done! I love this. Then, we'll work on the burgers. Preheat a large skillet to medium-high heat with some olive oil, preheat the broiler to high, then add all the spices to the ground turkey.\\\\
Whether you like your velvet crushed, vibrant or head-to-toe, there's really no denying the sheer luxe and elegance of this timeless textile. Not only is it super stylish, it can actually be so wearable for day-to-day wear. Yes, really! This year it's all about embracing fun gem-toned velvety pieces. Long gone are the days when velvet was solely associated with dark moody shades of navy and black. Below we've rounded up the most covetable velvet pieces on the high street right now. We're already coming up with outfit ideas! Are you completely obsessed or beyond bored of it?\\\\
Save up to \$8,086 on one of 1,258 Chrysler 200s near you. Find your perfect car with Edmunds expert and consumer car reviews, dealer reviews, car comparisons and pricing tools. We have 4,850,420. Research 2015 Chrysler 200 Sedan 4D 200C I4 prices, used values \& 200 Sedan 4D 200C I4 pricing, specs and more. Many years ago, we wrote about the stalling problem with the 2011 Chrysler 200, and believe it or not, we still receive an occasional call regarding the problem.However, a much larger issue has monopolized the phone lines as of late 2015 Chrysler 200 transmission problems leaving drivers with check engine lights, harsh shifting, and the occasional loss of power. The 2015 Chrysler 200 can fetch a premium for its style and its horsepower--but rear-seat room and handling are better bargains elsewhere. Find out why the 2015 Chrysler 200 is rated 8.4 by The. Don't know where to find the perfect rims for your 2015 Chrysler 200 CARiD.com stores a massive selection of 2015 Chrysler 200 wheels offered in myriads of design and finish options, including chrome, black, silver, and so much more.}
\end{tcolorbox}

\clearpage

\subsection*{Medium Style}
\begin{tcolorbox}[colback=gray!20, colframe=gray!75, rounded corners, sharp corners=northeast, sharp corners=southwest]
\texttt{In the initial stage of the survey on work-related stress, participation is optional, and all responses will be kept confidential. It is important to note that only those with work experience, whether part-time or full-time, will be able to accurately respond to certain questions. To complete the questionnaire, please use the following link, which is available in multiple languages.\\\\
Frozen burgers are not inherently flawed, but the meat seasonings are a crucial element that elevates the flavor of the dish. Ground turkey can be quite bland without the right blend of spices, so it is essential to add onion powder, chili powder, salt, pepper, and cayenne pepper to the mixture. Additionally, garlic and onion should be used in the mayonnaise, which is necessary for the burger's topping. To make the mayo, grate the garlic directly into the mixture, add a pinch of salt, and squeeze in lemon juice. Mix well. The mayonnaise is a delicious addition to the burger topping. Now, let's work on the burgers. Heat a large skillet over medium-high heat with some olive oil, and preheat the broiler to high. Add the seasonings to the ground turkey and mix well. The burgers should be cooked on the skillet for about 5 minutes on each side, then transferred to the broiler to cook for an additional 2-3 minutes. The burgers are ready to be assembled on buns, along with swiss cheese, lettuce, and onion. Tomatoes can be added as a topping, but they may overpower other flavors. It is up to the individual to decide if they want to include them in their burger. In conclusion, frozen burgers can be tasty, but the meat seasonings are essential to add depth and flavor to the dish. The mayonnaise and toppings can also be customized to suit individual preferences.\\\\
The luxurious and elegant nature of velvet cannot be denied, whether it is crushed, vibrant, or head-to-toe. This timeless textile is not only fashionable but also wearable for daily wear. In recent years, there has been a shift towards embracing fun gem-toned velvety pieces, breaking away from the traditional association of velvet with dark moody shades of navy and black. Below are some of the most desirable velvet pieces available on the high street today, and we are already envisioning outfits around them. Are you completely enamored with velvet or simply beyond bored with it?\\\\
Discover savings up to \$8,086 on one of 1,258 Chrysler 200s near you. Get expert and consumer car reviews, dealer reviews, car comparisons, and pricing tools from Edmunds. We have 4,850,420 listings for 2015 Chrysler 200 Sedan 4D 200C I4 prices, used values, and more. While we wrote about the stalling problem with the 2011 Chrysler 200, we still receive occasional calls regarding the issue. However, a more significant issue has taken over the phone lines recently - 2015 Chrysler 200 transmission problems that leave drivers with check engine lights, harsh shifting, and occasional loss of power. Despite the premium price for its style and horsepower, rear-seat room and handling are better bargains elsewhere. The 2015 Chrysler 200 is rated 8.4 by The Car Connection. If you're looking for the perfect rims for your 2015 Chrysler 200 CARiD.com offers a vast selection of 2015 Chrysler 200 wheels in various design and finish options, including chrome, black, silver, and more}
\end{tcolorbox}

\clearpage

\subsection*{Q/A Style}
\begin{tcolorbox}[colback=gray!20, colframe=gray!75, rounded corners, sharp corners=northeast, sharp corners=southwest]
\texttt{Question: Is participating in the stress at work survey mandatory? Answer: No, it is voluntary. Question: Will all answers be saved anonymously? Answer: Yes, they will. Question: Who is eligible to take the questionnaire? Answer: Only individuals with work experience, whether part-time or full-time. Question: What happens if someone is not eligible to take the questionnaire? Answer: They will not be able to answer some of the questions. Question: Can you provide a link to all language versions of the questionnaire? Answer: Yes, here is the link.\\\\
Question: What makes these frozen burgers so good? Answer: The meat seasonings are strong and spicy, and they give the burgers a good flavor. Question: Do you think ground turkey can be flavorless? Answer: Yes, ground turkey can be flavorless, but adding spices and seasonings can help. Question: What do you need to make these turkey burgers? Answer: Ground turkey, onion powder, chili powder, salt, pepper, and cayenne pepper for the burgers. Then the mayo takes garlic and onion. Buns, swiss cheese, lettuce, and onion are also needed. Question: What about toppings? Answer: Add tomatoes if you'd like to your array of toppings. Question: How do you make the mayo? Answer: Grate the garlic directly into the mayo, add a pinch of salt, and squeeze in the lemon juice. Stir. Done! Question: How do you work on the burgers? Answer: Preheat a large skillet to medium-high heat with some olive oil, preheat the broiler to high, then add all the spices to the ground turkey.\\\\
Question: What do you think of velvet? Answer: Whether you like your velvet crushed, vibrant or head-to-toe, there's really no denying the sheer luxe and elegance of this timeless textile. Question: Is velvet super stylish? Answer: Yes, really! This year it's all about embracing fun gem-toned velvety pieces. Question: Was velvet solely associated with dark moody shades of navy and black? Answer: Long gone are the days when velvet was solely associated with dark moody shades of navy and black. Question: Do you have any covetable velvet pieces on the high street right now? Answer: Below we've rounded up the most covetable velvet pieces on the high street right now. Question: Are you completely obsessed or beyond bored of it? Answer: We're already coming up with outfit ideas!\\\\
Question: How can I save money on a Chrysler 200? Answer: Save up to \$8,086 on one of 1,258 Chrysler 200s near you. Find your perfect car with Edmunds expert and consumer car reviews, dealer reviews, car comparisons and pricing tools. Question: What are the problems with the 2015 Chrysler 200? Answer: We have 4,850,420. Research 2015 Chrysler 200 Sedan 4D 200C I4 prices, used values \& 200 Sedan 4D 200C I4 pricing, specs and more. Many years ago, we wrote about the stalling problem with the 2011 Chrysler 200, and believe it or not, we still receive an occasional call regarding the problem. However, a much larger issue has monopolized the phone lines as of late 2015 Chrysler 200 transmission problems leaving drivers with check engine lights, harsh shifting, and the occasional loss of power. Question: What are the benefits of buying a 2015 Chrysler 200? Answer: The 2015 Chrysler 200 can fetch a premium for its style and its horsepower--but rear-seat room and handling are better bargains elsewhere. Question: How is the 2015 Chrysler 200 rated? Answer: It's rated 8.4 by The. Question: Where can I find the perfect rims for my 2015 Chrysler 200? Answer: CARiD.com stores a massive selection of 2015 Chrysler 200 wheels offered in myriads of design and finish options, including chrome, black, silver, and so much more.}
\end{tcolorbox}

\end{document}